\documentclass{article}
\usepackage{graphicx} 
\usepackage{amsmath}
\usepackage{amssymb}
\usepackage{booktabs}
\usepackage{makecell} 
\usepackage{multirow} 
\usepackage[a4paper,margin=1in]{geometry}
\usepackage{comment} 
\usepackage{array, tabularx}
\usepackage{longtable}

\title{Review: Latent representation models in neuroimaging}
\author{
    C. Vázquez-García$^{1}$, 
    F. J. Martinez-Murcia$^{1}$, 
    F. Segovia Román$^{1}$, 
    Juan M. Górriz$^{1}$
    }

\date{December 2024}

\begin{document}

\maketitle

\begin{center}
$^1$ Department of Signal Theory, Telematics and Communications, University of Granada, Spain
\end{center}

\begin{abstract}
    Neuroimaging data, particularly from techniques like MRI or PET, offer rich but complex information about brain structure and activity. To manage this complexity, latent representation models --such as Autoencoders, Generative Adversarial Networks (GANs), and Latent Diffusion Models (LDMs)-- are increasingly applied. These models are designed to reduce high-dimensional neuroimaging data to lower-dimensional latent spaces, where key patterns and variations related to brain function can be identified. By modelling these latent spaces, researchers hope to gain insights into the biology and function of the brain, including how its structure changes with age or disease, or how it encodes sensory information, predicts and adapts to new inputs. This review discusses how these models are used for clinical applications, like disease diagnosis and progression monitoring, but also for exploring fundamental brain mechanisms such as active inference and predictive coding. These approaches provide a powerful tool for both understanding and simulating the brain’s complex computational tasks, potentially advancing our knowledge of cognition, perception, and neural disorders.
\end{abstract}

\section{Introduction}

Neurodegenerative diseases (NDDs), such as Alzheimer (AD) or Parkinson (PD) are amongst the most prevalent in the world, affecting over 29.8 million and 6.2 million people respectively. In the last years, the prevalence of these diseases has escalated drastically, with more than a twofold increase from 1990 to 2015. This increase has led to a decline in quality of life as the average life expectancy gets larger. In addition, there is a large spectrum of neurological disorders that affect millions of people all over the world across all ages, such as Schizophrenia (SZ), ADHD, Autism or Brain tumors, hindering their daily lives. These diseases do not only have a severe effect on the healthcare system, but they also have an impact on the economy. Therefore, the need for effective clinical therapies is becoming increasingly urgent.

Currently, most therapeutic approaches prioritize improving patients' quality of life and alleviating symptoms, rather than directly addressing the underlying causes of the disease. Although these treatments may offer temporary relief for certain individuals, their efficacy is often case-dependent and tends to diminish over time. Moreover, they are unable to halt or reverse disease progression, as clinical trials generally begin many years after the neurochemical processes driving the condition have already been set in motion.

Research in neuroscience has shown that neuropathological processes of NDDs such as AD or PD begin years, even decades, before the first symptoms appear. This presymptomatic stage of the diseases opens a window of opportunity to implement preventive therapies and treatments. In this context, Disease-Modifying Therapies (DMTs) have emerged as a basis for developing new therapeutic techniques. These therapies aim to modify the pathological processes underlying neurological diseases by targeting the mechanisms that drive their progression . Without such a comprehension, the identification of the right mechanisms to target may be misinterpreted. For instance, most pharmacological trials for AD have been developed to decrease the levels of Amyloid-beta (A$\beta$) aggregates and plaques by using drugs. Nevertheless, these trials have not been successful in retrieving the phenotypes of the disease, compelling many researchers and funding bodies to shift the focus to another potential causes. On the other hand, non-degenerative diseases, such as neurological disorders, even though they might not be as dramatic as NDDs, they also hinder severely the lives of the patients in most cases. Alike NDDs, the mechanisms underlying these disorders are still not well understood. Moreover, the vast heterogeneity of the spectrum of these disorders greatly difficult the progress in research.

Beyond clinical practice, there is a growing interest in understanding the fundamental mechanisms of brain function. Research has revealed that the brain operates as a higly dynamic system capable of encoding, predicting, and adapting to environmental inputs. These processes, including sensory encoding, predictive coding, and active inference, reflect the brain's ability to generate internal representations that optimize perception and behavior. However, unraveling these mechanisms remains a challenge due to the high-dimensional nature of neuroimaging data, which captures both brain structure and function but is difficult to interpret effectively. 

In this context, research has focused on designing techniques that allow to interpret and capture relevant insights from these non-invasive neuroimaging techniques. In the last years a variety of models and techniques have been developed to analyze clinical data, among them latente generative models have emerged as powerful tools for addressing these challenges. This review aims to provide a comprehensive overview of how latent generative models are applied in neuroimaging. We explore their role in clinical tasks, such as disease diagnosis, progression monitoring, and harmonization of multi-site data, as well as their contributions to fundamental neuroscience, including the study of brain networks, cognitive processes, and sensory representations. By bridging clinical and theoretical research, these models offer new opportunities to advance out understanding of brain function and dysfunction, ultimately paving the way for improved diagnosis, therapies, and insights into the brain's remarkable computational capabilities.

\section{An opportunity: latent representations}
Medical images are commonplace in hospitals and laboratories worldwide. In neuroimaging, these images are extensively used for their non-invasive nature and the valuable insights they provide into the structure and functionality of the brain. In the context of neurodegenerative diseases (NDDs) and neurological disorders, these neuroimaging techniques play a critical role in monitoring disease progression and assisting clinical decisions. Despite their utility, neuroimaging data can be highly complex, requiring both technological expertise and an understanding of the diseases being studied. Moreover, the high dimensionality of these data, when combined with other biomarkers, presents significant challenges for modeling and interpretation.

One promising approach to tackling these challenges comes from the \textit{manifold hypothesis}, which suggests that real high-dimensional data tend to lie on or near a lower-dimensional manifold. This means that even though the data may have a large number of dimensions, much of it is constrained to a more compact, lower-dimensional structure. This hypothesis, first explored in mathematics and later applied to various fields like psychology and neuroscience, proposes that the high-dimensional data we collect, including neuroimaging data, can be better understood by uncovering this underlying manifold structure.

\begin{figure}[ht] 
    \centering 
    \includegraphics[width=0.8\textwidth]{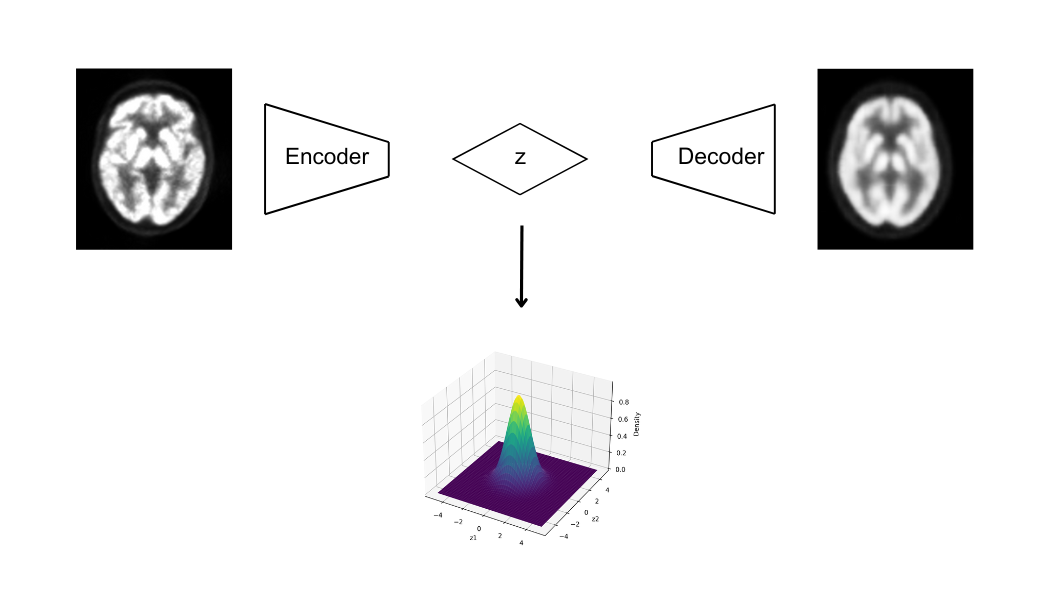} 
    \caption{Simplified scheme of the latent representation. High-dimensional neuroimaging scans are processed through a model capable of computing the probability distribution of a lower-dimensional manifold that captures relevant patterns in our dataset, whether the model is implicit or explicit.}
    \label{latent_reps} 
\end{figure}

The \textit{manifold hypothesis} has become particularly influential in the analysis of neural data. In psychology, for instance, it has been suggested that the brain encodes information in a lower-dimensional space, which allows it to process complex sensory input efficiently \cite{chung2021neural, lindsay2022uncovering, jazayeri2021interpreting}.
One usual problem of high-dimensional data, is the \textit{curse of dimensionality} \cite{friedman1997bias}, which suggests that when the number of features (e.g., voxels in an image) exceeds the number of samples, traditional methods may fail to provide reliable models. For example, neuroimaging volumes of dimensions $90 \times 110 \times 90$ can result in hundreds of thousands of parameters, but we typically have only a few thousand samples, making our data space very sparse. Thus, uncovering the latent manifolds that are believed to exist in high-dimensional neuroimaging data becomes essential for understanding the brain's complexities.

To explore the idea of the manifold hypothesis in data science, various mathematical and computational methods have been proposed. Techniques such as Principal Component Analysis (PCA) and Multi-Dimensional Scaling (MDS) have long been used to reduce the dimensionality of data and identify its latent structure. However, these techniques rely on linear assumptions and may fail to capture complex, non-linear relationships within the data. For instance, PCA and MDS use Euclidean distances to define proximity between data points, but this can be misleading in the presence of non-linear manifold structures \cite{tenenbaum2000global}.

In this context, Locally Linear Embedding (LLE) \cite{roweis2000nonlinear} emerged as a method that attempts to capture non-linear structures by modeling data based on local relationships between neighboring points. While elegant, these techniques are sensitive to noise and may perform poorly if the data does not satisfy the assumptions necessary for their successful application.

The rise of Deep Neural Networks (DNNs) has brought new methods for modeling complex latent manifolds. Unlike linear methods, DNNs can capture non-linear relationships within data, making them ideal for high-dimensional neuroimaging data. Models like Autoencoders (AEs), Generative Adversarial Networks (GANs), and Latent Diffusion Models (LDMs) are particularly well-suited for this task, as they learn to represent data in lower-dimensional latent spaces through non-linear transformations. These approaches have opened up new avenues for understanding the underlying patterns in neuroimaging data, contributing not only to clinical diagnosis but also to our understanding of the brain’s encoding processes.

\section{Latent Generative Models}

Deep learning models have proven highly effective for deriving latent representations of complex, high-dimensional, non-linear data in neuroimaging, as highlighted in the previous section. This is typically achieved through the use of latent generative models, which are designed to learn compact representations of input data while simultaneously generating new samples that adhere to the true underlying data distribution. These models excel at capturing the essential features and variations within the data, enabling a robust characterization of its distribution. In this section, we delve into the foundational principles underlying the most widely employed latent generative models, setting the stage for subsequent sections where their applications in neuroimaging will be examined in detail.

\subsection{The (Variational) Autoencoder}
An Autoencoder (AE) is a neural network that learns to encode input data into a lower-dimensional latent space and then reconstruct it back to its original form. It is commonly used for dimensionality reduction, feature extraction, and denoising. The AE consists of two main components: an encoder, which compresses the input data into the latent representation, and a decoder, which reconstructs the data from this compressed representation.

While AEs are effective at capturing key features of the data, their latent representation is learned directly from the data without additional constraints. This means that the structure of the latent space may not be well-organized or interpretable, and it might not effectively capture the variability within the dataset. For example, in the MNIST dataset of handwritten digits, each digit has unique variations depending on the writer's style (e.g., different stroke thicknesses or curvatures for the digit '2'). An AE might learn a representation for each sample that captures its unique details, but these representations might not generalize well to unseen variations.

\begin{figure}[ht] 
    \centering 
    \includegraphics[width=0.8\textwidth]{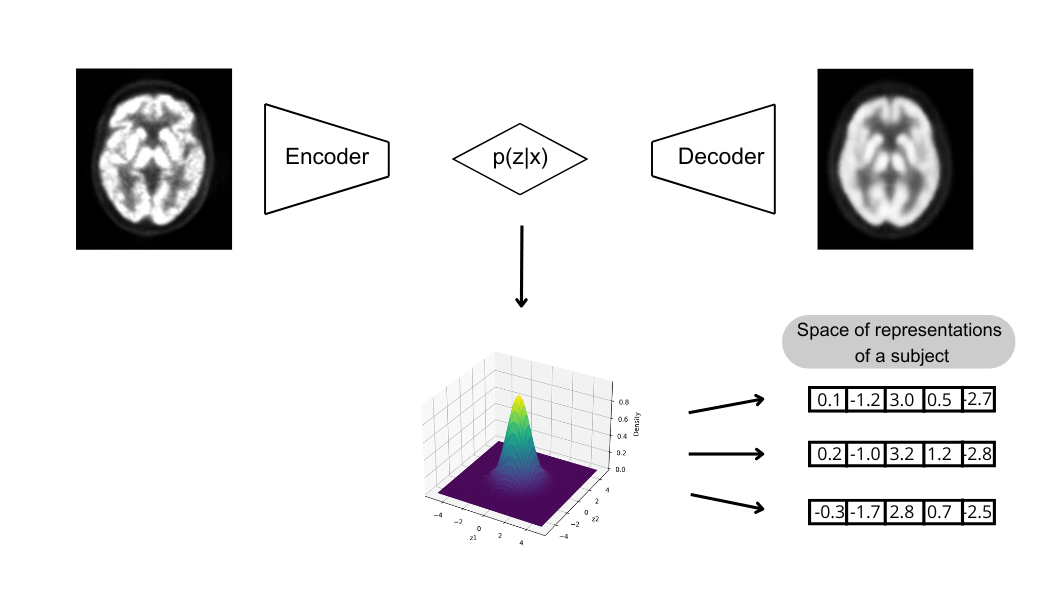} 
    \caption{Simplified scheme of the VAE. This explicit model captures the priori distribution that encodes high-dimensional data into the latent manifold, allowing to sample and access the latent variables of the model.}
    \label{vae_scheme} 
\end{figure}

Similarly, when applied to datasets like magnetic resonance images (MRI) of healthy subjects and subjects with neurodegenerative diseases (NDDs), an AE might capture general brain patterns but fail to account for inter-subject variability. For instance, two patients with Alzheimer's disease (AD) might exhibit different patterns of brain degeneration, one might show severe hippocampal atrophy, while another might have more widespread degeneration. An AE may not organize the latent space in a way that reflects these distinct pathological variations, limiting its ability to model and generalize the variability in the data.

This problem was tackled by Kingma in its work \cite{kingma2013auto}, where he introduced the Variational Autoencoder (VAE), which is a modification of the usual AE, based on Bayesian Inference.  This generative model, instead of learning a fixed representation, VAEs learn a probabilistic distribution of the latent representations. In this work, Kingma proposed a way to learn the intractable posterior distribution that encode the real data into the latent representation. The idea is that neuroimages of a certain disease $X$ are produced by an underlying pathological process $Z$ that we do not know. Mathematically, this defines a joint probabilistic model known as generative model $p(x,z) = p(x)p(z|x)$, where $p(x)$ is the marginal likelihood of the neuroimaging data, and the posterior distribution $p(z|x)$ models how this data is encoded into the underlying latent representation. In practice, these probabilities are intractable due to the complexity, and so a variational inference (VI) model $q_{\phi}(z|x)$ is introduced to substitute the real posterior distribution, which is parameterized by a deep neural network known as the encoder. Using this VI approach we can obtain a lower bound of the marginal loglikelihood of data $p(x)$:
\begin{equation}\label{ELBO}
    \log p(x) \geq E_{\phi}[\log p(x|z)] - D_{KL}(q_{\phi}(z|x)||p(z)).
\end{equation}
The first term can be interpreted as a reconstruction error, which tries to minimize the structural difference between real and synthetic data, while the second term $D_{KL}$ is the Kullback-Leibler divergence, which measures the difference between the VI posterior distribution and the prior distribution of the latent variables $p(z)$. This ELBO loss is a competition between precise reconstruction of images and regularization of the latent space, i.e., capturing relevant variability of the data. If the KL divergence is very low the reconstruction will be very accurate but the variability of the latent space will be very small, which is undesirable.

The main advantage of the VAE-based model is that the information codified into the latent representation contains precious information about the real neuroimages, since they are trained to mimic the true codifying distribution, however, the reconstructions are usually very blurry compared to other DL models. This is to be expected because we can see in the ELBO equation \eqref{ELBO} that reconstruction detail is sacrificed for the sake of a meaningful latent representation.

\subsection{Latent Diffusion Models}

Latent Diffusion Models (LDMs) are latent generative models that focus on transforming the input data into noise through a diffusion process and learn how to revert that process to generate high quality samples. The main idea of this process is to map the complex data into a latent representation, where the diffusion process is more efficient and easy to handle, in order to learn how to generate new synthetic data in this latent representation.
The LDM is divided in three main stages: i) encoding of the input data into a latent representation, which is usually performed with an encoding model, such as a AE, ii) forward noisy diffusion, and iii) reverse process.
During the forward diffusion, random gaussian noise is added to the input data. This is performed via a Markov process, meaning that each diffusion step depends only on the immediate previous step and not in the history of the process. At each step, a small amount of noise is added to the entire latent representation, such that the process is described by:
\begin{equation}
    q(z_t|z_{t-1}) = \mathcal{N}(z_t; \sqrt{1-\beta_t}z_{t-1}, \beta_t I).
\end{equation}
This distribution describes how the diffusion step $z_t$ is produced by the previous step $z_{t-1}$. This can be translated as, given $z_{t-1}$, $z_t$ is distributed as a normal distribution with mean $\sqrt{1-\beta_t}z_{t-1}$ and variance $\beta_t I$, where the term $\sqrt{1-\beta_t}$ controls how much of $z_{t-1}$ is conserved and $\beta_t$ controls how much noise is added. The entire diffusion process is simply the product of each diffusion step.

Once the data has turned into noise, the objective of the LDM is to train a generative model that learns how to reverse this process and retrieve the original representation $z_0$ from the final noisy step $z_t$. To do that, a UNet or transformer is commonly trained to estimate the noise $\epsilon_{\theta}(z_t,t)$. To do so, the goal is to minimize the difference between estimated noise and real noise added to the input:
\begin{equation}
    \mathcal{L}(\theta) = \mathbb{E}_{z_0,t}\left[ ||\epsilon_{\theta}(z_t,t) - \epsilon(z_t) ||^2 \right],
\end{equation}
where $\epsilon(z_t)$ is the real noise at step $t$ and $\epsilon_{\theta}(z_t,t)$ is the prediction of the model. Once the denoising process is finished we can reconstruct the synthetized neuroimage by using the decoder of the model we used to encode the images into the latent representation as $\hat{x} = Decoder(\hat{z})$. As we will see this model is commonly used in neuroimaging to produce high quality images from latent representations. 

\begin{figure}[ht] 
    \centering 
    \includegraphics[width=0.8\textwidth]{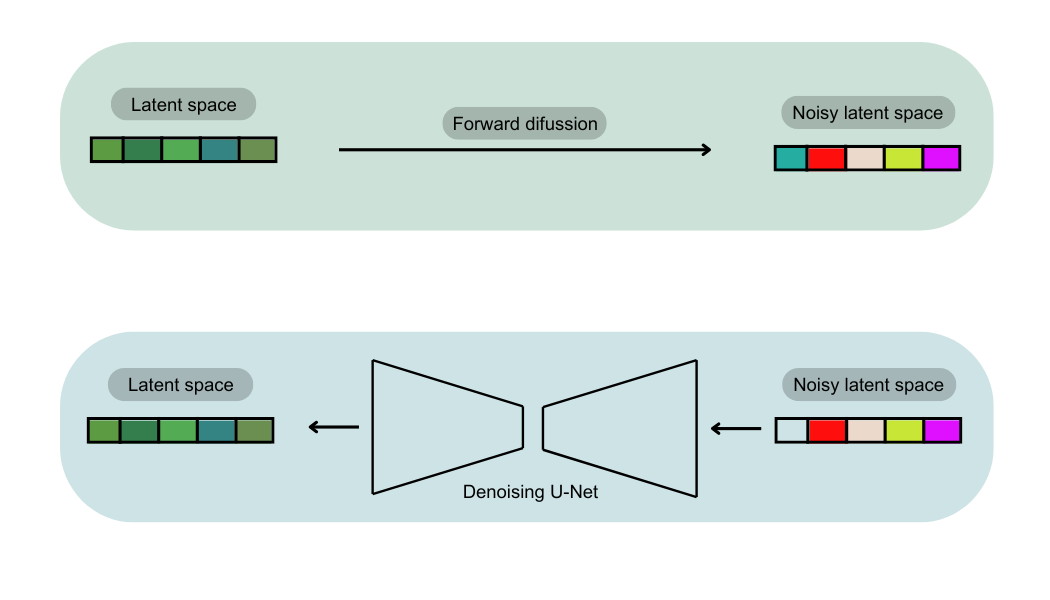} 
    \caption{Simplified scheme of the LDM. This latent generative model takes a latent space and generate a forward diffusion, introducing noise into this representations. Then, an inverse process learns to denoise the latent space to generate a reconstructed latent space using a denoising U-Net.}
    \label{LDM_scheme} 
\end{figure}

As we have discussed before, VAEs tend to reconstruct blurry images due to their nature, in order to obtain meaninful representations. If we combine a VAE model with a LDM we can exploit the rich representations learnt by the VAE and still generate high quality images. Notice that, unlike the VAE, LDMs do not learn how to capture relevant underlying information about neurological processes hidden in neuroimages or how to model their distributions. Instead, they need a previous step to obtain representations and then they make use of that information to generate new data. In that sense, LDMs can be thought as a tool to handle the rich information of the latent space, more than a method to extract them.

\subsection{Generative Adversarial Networks}

GANs were introduced by Goodfellow in 2014 \cite{goodfellow2014generative} as DNN based on two counterparts: a generator G and a discriminator D. The GAN can be thought as a competition between a generative model that is learning how to generate synthetic data that resembles the input, and discriminator that is trying to guess whether the received image is real o generated. The generator G takes a noise vector $z$ from a latent space and maps it into a data space of neuroimages, while the discriminator takes a sample $x$ which can be real or generated, and computes the probability of $x$ being real. Hence, it is formalised a a minmax problem, where the loss function is:
\begin{equation}\label{GAN_loss}
    \min_G \max_D V(D,G) =  \mathbb{E}_{x\sim p_{data}(x)}[\log D(x)] + \mathbb{E}_{z\sim p_z(x)}[\log(1-D(G(z)))].
\end{equation}

The goal is to learn the generator distribution $p_g$ over the data x. We can think of the model as a child who is trying to learn how to draw. The generator is the child and his task is to make the drawings look real. The discriminator is a teacher who is trying to guess whether the drawing is real or not (generated). The generator is trying to fool the discriminator and the discriminator is trying to improve in order not to be fooled. If we take a look at the loss function \eqref{GAN_loss} what we see is that the generator wants to minimize V(D,G) and the discriminator is trying to maximize it. The first term of that equation assures that the discriminator classifies the real samples as real, while the second term assures that the generator produces better images.

\begin{figure}[ht] 
    \centering 
    \includegraphics[width=0.8\textwidth]{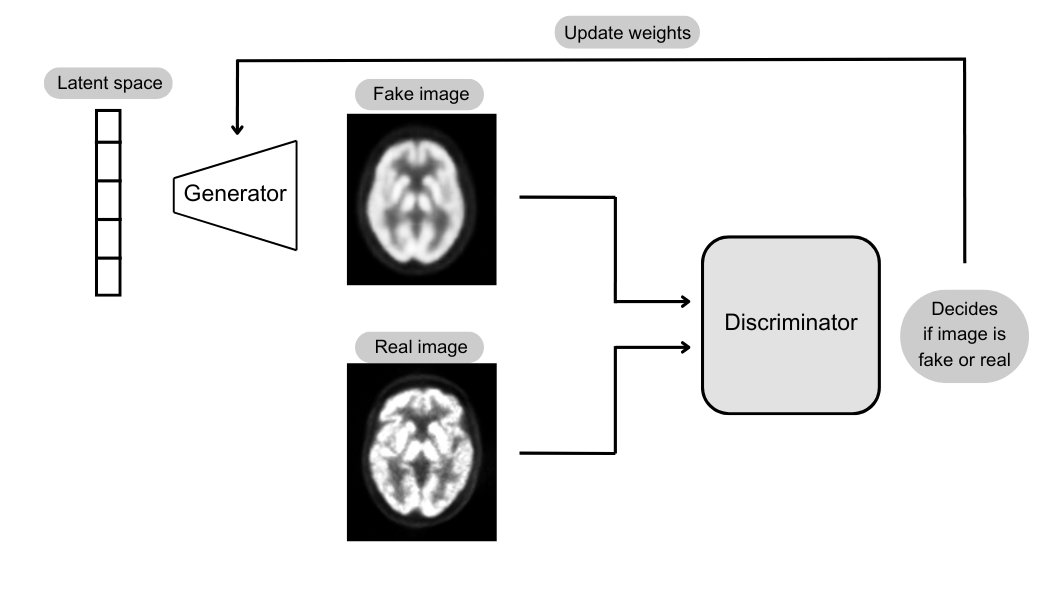} 
    \caption{Simplified scheme of the GAN model. A latent space is fed to a generator that produces fake reconstructed images. Then, a discriminator learns to distinguish between real and fake images, updating the weights of the model according to the loss function}
    \label{GAN_scheme} 
\end{figure}

Now, we have to take into account that, even though GAN models can learn a relationship between latent representations and real data, this latent space is not structured or explicitly defined. While VAEs learn an explicit latent representation that we can interpret and exploit for different analyses, GANs learn these relationships implicitly. Specifically, GANs optimize their loss function through the adversarial ``game'' between the generator and the discriminator, without approximating a probabilistic distribution explicitly, as VAEs do. However, the representation of GANs is implicit and inaccessible for analysis, making it closer to implicit models like Latent Diffusion Models (LDMs), useful for high-quality generation or synthesis.

\section{State-of-the-art}

In this section we will present the most relevant lines of research in obtainment, exploitation, and analysis of latent representations through latent generative models in neuroimaging. Here, we will explore how researches have been using the bayesian inference framework as a mean for understanding the internal cognitive processes and neural connections in the brain. Additionally, we will also explore how these latent representations are used to create models that aim to aid typical clinical practice problems such as the inter-center noise due to the different use of scans and methodologies of each institution, as well as the translation of low resolution neuroimages into higher resolutions, like 7T MRI. Moreover, we will see how these latent representations are able to capture longitudinal patterns of the brain, allowing us to model brain aging or classify different neurodegenerative diseases or disorders, such as Alzheimer or Schizophrenia. Furthermore, we will discuss how functional brain networks can be extracted using latent spaces, providing us with relevant insights of how these brain networks work for subjects with certain disorders, such as ADHD. Next, we will see how these latent representations are useful for integrating different types of data, an open problem in medicine that grants us with great insight of diseases. Finally, we will also see how we can both synthesize images in order to augment our datasets or even reconstruct visual information through fMRI maps, as well as other works based on latent representations.

\subsection{The Brain as an inference machine}

As we have discussed, generative models are bioinspired by the encoding and decoding mechanisms of the brain. However, these models are not just powerful tools to analyze or generate data, but they also give us an opportunity to understand the most abstract and fundamental processes of the brain. 

Traditionally, it was thought that the brain receives stimuli from its surroundings and processes them to form perceptions. However, we now know from psychology that perception is not a passive process. Instead, the brain actively generates inferences about the environment based on prior knowledge and expectations. This active inference means that the brain not only perceives but also predicts what is going to happen next, using previous knowledge and past experiences to predict future outcomes. This actually resembles the usual Bayesian Inference that we are familiar with, where we could consider that the brain is producing constant beliefs about the world and updating those beliefs according to the sensory information following the Bayes' rule:
\begin{equation}
    p(z|x) = \frac{p(x|z)p(z)}{p(x)},
\end{equation}
where $p(z)$ is the belief and $p(x|z)$ is the likelihood that the observation $x$ fits the belief $z$. 
The brain is, without a doubt, a physical object, and thus there must exist a mathematical formalism to describe it. While this formalism may be highly complex, Bayesian inference, though not the ultimate answer, undoubtedly brings us a step closer to understanding how neural encoding, perception, and cognition operate. In this context, several researchers have leveraged the powerful Bayesian framework to gain deeper insights into how inference and representations are generated within the brain.

For instance, authors in \cite{friston2001generative} discuss the relationship between generative models, brain function and neuroimaging. They focus on how functional specialization of the brain depends on the integration between population of neurons. They use generative models to explain the paper of feedback connections among neurons allowing cortical regions to reconfigure dynamically depending on the context. The authors explore the idea that forward and backward connections in deep learning models are very similar to the bottom-up and top-down connections of neurons. Bottom-up connections refer to the flow of information from lower levels (sensory areas that receive stimuli) to higher levels (areas that process more complex information). This represents the data-driven pathway, where stimuli are progressively transformed into perceptions. In top-down the information flows from higher to lower areas, in which case, higher areas of the brain generate predictions about what they expect to receive and send these predictions to lower areas. These predictions act as a guide or context for a better interpretation of the received sensory information. If the prediction does not coincide with the input information the neuron population would produce an error that error is used to adjust the predictions in the higher areas until these predictions coincide with the perceived reality. The error disappears once there is a consensus between prediction and reality.

 The main idea of the authors is that processing within the brain does not only take into account information coming from stimuli (bottom-up) but also from the predictions created by the active inference of the brain. Moreover, this means that what we perceive is not only a product of the stimuli but it is also influenced by the constant context and prediction generated by the brain based on the previous experiences and environment. The authors in this paper argue that the brain can be conceptualized as a system that is constantly generating predictions about the external world, adjusting these predictions according to the feedback provided by the external stimuli. These generative models allow the brain to construct an internal representation of reality that helps to optimize its prediction ability and reaction to stimuli.

Here, they also propose that this theory can be experimentally validated using neuroimaging techniques. First, they consider a task consisting of saying "yes" when the subjects see a recognisable object, and "yes" when they see an unrecognisable non-object. The idea is to see how recognisable objects are perceived in the brain by specific areas related to visual object recognition, while abstract, unrecognisable objects activate different areas. Using this comparison they can isolate which brain areas are actually related to object recognition. Now, they consider a factorial design, in which an additional component is added to the recognition task. Imagine the same object vs non-object recognition task but, instead of saying "yes" the subjects have to name the object (if it is a recognisable object) or name the colour of the non-object. In this task, there is an additional name retrieval component. This component modifies the brain response that is observed in the previous "yes" task, meaning that a new cognitive process is taking place (name retrieval, in this case). This task allows us to see how different areas interact between them. The main idea behind this experiment is that the context of the additional component influences the processing of object recognition. The name retrieval component adds a complexity layer on top of the visual recognition task. When the subject has to name the object, the brain predicts which information should be available, which modulates the perception of the object. The brain is not passively reacting to the bottom-up information, but it is also expecting a name and creating its own predictions. The brain does not simply react passively to bottom-up information; it also anticipates and generates its own predictions. Furthermore, the literature has demonstrated that top-down information flows play a significant role in this process. For example, as \cite{friston2003learning, friston2005theory} mention, mention, the perceptual encoding of "moving quickly" is known to occur in the visual area V5, while the visual area V2 is responsible for processing basic shapes and borders. It has been observed that predictions generated in V5 are sent back to V2, influencing the processing of visual information to what is expected to receive. Moreover, authors in \cite{kao2021optimal} have found, in motor preparation, that the brain is able to find optimal preparation subspaces that are closer to motor response. In addition, they discuss that this optimal anticipatory control of future movements requires a form of internal feedback which is implemented through a thalamic-cortical loop. This aligns to the idea that there is a forward and backward propagation and the idea of an internal prediction. 

Moreover, the authors discuss in a review on the impact of computational neuroscience on neuroimaging \cite{friston2010computational} how latent generative models are essential for understanding latent representations in neuroimaging. They note that these models provide access to latent computational variables, which represent mental states inferred during tasks. This is compared to the brain's ability to infer the mental states of others, a process extensively studied in psychology known as the Theory of Mind \cite{premack1978does}. The authors claim that, by using generative models, the beliefs of a subject over another individual during a game can be estimated, which is translated to neural correlations in fMRI. This approach allows mapping of computational variables to neural representations. To exemplify this generative brain idea they use the paradigm of the Theory of Mind, by using the cooperative game experiment. In this experiment, the participants play a cooperative known as "stag hunt". In this game, the subjects have to either hunt smaller prey on their own (rabbits) or cooperate with other players to hunt larger prey (stag). When we have two or more players, there is a joint state space of all players, representing their mental representations, that is, each representation can not be understood individually but they are affected by the presence of other players. This gives raise to a difficult problem known as infinite regression. If I want to evaluate your state, I need to know how you evaluate my state, for which you need to know how I evaluate your state evaluating my state, and so on. This problem can be addressed by assuming that this inference is not infinite but there is some limit or upper bound referred as the level of sophistication. In the stag game experiment the subjects were told that they were playing with other participants, but instead they were playing against a computer that could simulate different levels of sophistication (levels of inference over the other players' mental states). The computer could act as a player that had a lot of understanding of the game (high sophistication) or very little understanding (low sophistication). In this experiment the level of sophistication is a latent variable. In each round of the game the subjects had to make decisions based on their beliefs about the opponents. These decisions were used to update the sophistication of the opponent, which highly resembles Bayes' rule. The subject forms a priori distribution over the levels of sophistication of its opponent and, through observation during the game, they update such beliefs, creating a posterior distribution over the sophistication. Once the estimations have been obtained, they can be used as a function of the stimulus in a fMRI analysis to identify neural correlations associated to the sophistication representation and the uncertainty of the subject. In this paper the authors show experimental validation showing how the computed posterior distributions match the levels of sophistication simulated by the the computed as the game advances. They also show evidence that suggest functional specialization for inference on the sophistication \cite{yoshida2010neural}.

On the other hand, the author in \cite{gershman2019generative} proposes an alternative approach of this brain inference paradigm based on implicit density models. In this work they try to explain how the usual Bayes' rule, although  very powerful, can not explain certain problems like why do certain illusions feel so real. They mention the Ramachandran and Hirstein (1997)  experience about wallpaper in a bathroom \cite{ramachandran1997three}. When gazing at a wallpaper in a bathroom we subjectively experience the wallpaper in our periphery with high fidelity even though we perceive it with low fidelity. However, the wallpaper behind us is not experienced. We infer with high confidence that there is wallpaper behind us even though we have no experience of seeing it. Similarly, we know that there is one blind stop in our eye due to the optic nerve, where light is not processed. However, we are not aware of this effect because the brain is able to fill in the missing information by inferring. In this paper, the author argues that this problem can be issued by interpreting Bayes' rule from a different perspective. Instead of thinking of an explicit latent representation, where we can directly access the latent variables, they use an implicit model, where a "generator" draws samples from a generative model which are then fed, along with real sensory data, to a "discriminator" that tries to tell apart which samples are real and which are fake. As we discussed in the latent generative model section, this describes a GAN model, as opposed to an explicit model (e.g., VAE). They mention that, in this context, if the visual system plays the role of the generator and our perception plays the role of the discriminator, then the illusions can be understood as either the visual system reporting things that are not there (or failing to report things that actually are there), or the perception failing to discriminate between real experiences and false ones. As the author indicate, a dysfunctional generator (visual system) will produce abnormal content, while a dysfunctional discriminator (our perceptual experience) will report fake content as real.

The author then try to use this generative approach to explain the previous wallpaper experience. The generator produces a realistic simulation that the discriminator has to accept as either real or fake. In this context, perceptual experience hinges on whether the discriminator deems it "real." For instance, peripheral vision or the blind spot might be accepted as real, even though the brain does not attempt to reconstruct objects outside our visual field. While the generator cannot simulate what lies behind us with accuracy, it can create plausible reconstructions of the blind spot or peripheral vision through interpolation. The inference of something like wallpaper behind us is shaped by prior experiences rather than by direct visual reconstruction, distinguishing it from other scenarios.

Additionally, the author discusses how the GAN model could be applied to the brain and how the different components would correlate with the neural components. Neuroimaging studies have proven that the median anterior prefrontal cortex is implicated in discriminating whether a visual object has previously been seen or if it is a product of imagination \cite{kensinger2006neural}, suggesting that it helps to control the fidelity of perceived experiences. Moreover, certain disorders like schizophrenia have shown reduced activation of this area during reality monitoring tasks, which may explain why some people experience hallucinations or are not able to tell apart between real and imagined experiences. On the other hand, the generator model is very similar to the feedback mechanisms of the brain cortex, as we have previously discussed in other works \cite{friston2001generative}. The proposed adversarial model suggests that the brain might implement generative models through probabilistic population code, meaning that probabilistic representations of perception are distributed through a population of neurons, where activity of these populations reflect the probability of different perceptual beliefs.

Finally, the author explain the process of delusion formation using the adversarial brain paradigm. In this context, there is a two-factor \cite{coltheart2010abductive} failure, in which both generator and discriminator malfunction. The abnormal content from the defective generator (which is usually easily reported as fake by the discriminator) can be accepted because the discriminator is also defective, giving rise to delusions.

\subsection{Image-to-Image translation $\&$ Harmonization}

One open problem in the field of neuroscience analysis is that every research center, institution or collaboration, aiming to gather subject data in many modalities (ranging from neuroimaging scans such as MRI-T1, MRI-T2, PET, CT, etc. to other clinical data such as $A\beta$ amyloid, tau or even genetics), use their own measures and techniques to register such information. This means that each project chooses which biomarkers are going to be measured and the specifics of the scans used to obtain images. Even amongst a certain project there are usually differences in the scans used. This produces a tremendous problem in the data analysis field, since it hinders the possibility of validation. Any model that aims to produce insights about a certain disease must be applicable to any database in order to validate its results. However, in the field of neuroimaging, there are certain patterns that differentiate images that are taken using different methodologies, which is translated as inter-center noise. For example, if we try to train a VAE using images from two different databases, this inter-center noise will severely affect the coding of our latent representation, which will be noisy too, as a consequence. 
In order to address this issue, researchers have proposed different methods and techniques using generative models to isolate this inter-center noise and eliminate it from the images in order to obtain a harmonized dataset from two or more datasets. These models exploit the fact the the noise is codified in the latent space to isolate it. The. authors in \cite{lu2022image} discuss the possibility of domain image translation as a solution for multi-modal image co-registration, a process crucial to align images obtained from different modalities. This work suggests that translation techniques can help the transformation of images from different modalities in a a common representation space, where latent representations could be used to capture relevant features of images. However, these techniques do not only work for modality translation (e.g. from PET to CT), but also for any kind of domain translation (e.g. inter-site translation).

For instance, the authors in \cite{yang2020mri} use a simple GAN model based on a generator and a discriminator. Here, latent representations obtained through the generator are compared to real images from the target modality using a discriminator. The network is trained using adversarial loss and L1 loss that guarantees coherence between the original modality and the target modality by using low-level features along with high-level features. To validate this model they evaluated the performance of translation from different modalities, T1-T2, T2-T1, T1-T2flair, T2-T2flair, PD-T2 (proton density), T2-PD. Using the MAE, PSNR, MI (mutual information) and SSIM performance metrics, they find that the full model (GAN+L1) outperforms the traditional methods such as Random Forest (RF) and CA-GA (contextual GAN) in most metrics and modalities. Qualitatively, they see that the combination of adversarial and L1 losses produces clear images, less blurry than the other methods. Furthermore, the authors employ their model to assess the cross-modality registration task, utilizing real T2 images and generated T1 images. They create two sets of subjects: one set, consisting of fixed subjects with T2 images and generated T1 images, serves as a reference, while the second set, comprising moving subjects, is used to align their T2 images with those of the reference set and their real T1 images with the generated T1 images. Each registration produces deformation fields, which are subsequently combined to generate a fused deformation field that integrates information from both processes. The registration performance is evaluated using the Dice coefficient and the Distance Between Corresponding Landmarks (DBCL) in brain structures. The results demonstrate that the inclusion of the generated modalities contributes to an improvement in the Dice score and a reduction in the distance, indicating a more accurate registration compared to using only real modalities. Finally, they used the model for the segmentation task in which they found that the generated modality adds additional information to improve the segmentation, compared to using a single modality. They evaluated a tumor segmentation task in the Brats2015 dataset, obtaining relevant performance metrics.

Authors in \cite{armanious2020medgan} proposed a similar but sophisticated model named MedGAN. Instead of simply using adversarial and L1 losses, this model combines adversarial and non-adversarial (content and style) losses to translate neuroimages from one domain to another, specifically PET-CT translation. The model consists of a CasNet generator which processes the input (PET) images to translate them to CT images through several coding-decoding U-blocks. These blocks extract high-level features of the images. In order to assure that the texture and structural features of the transformed input image coincide to the target domain features they use style-content losses (non-adversarial loss). The content loss makes sure that the anatomical content of the image is preserved, while the style loss translates the visual features of the image from origin to target domain. The content loss obtains the representation of the real CT image used to train and the generated CT to compare them. Finally, a discriminator (adversarial loss) trains the model to tell apart real from generated input. In order to validate their model they perform ablation studies and compare the real and generated data using several performance metrics such as SSIM, PSNR, MSE, UQI (universal quality index) and LPIPS (learned perceptual image patch similarity). They obtain the best values of all metrics using the MedGAN architecture with every component (style loss, content loss, perceptual loss, and adversarial loss). They also compare these results with other state-of-art models such as pix2pic, ID-CGAN, Fila-sGAN and find that MedGAN outperforms them in most cases. This work reveals the relevance of both the anatomical and style latent representations of the images. These lower-dimensional manifolds are able to capture biological and visual properties of the data, allowing both a better performance and interpretability.

Authors in \cite{kim2024adaptive} use a LDM to translate 3D MRI across different modalities. Here, the latent representations are used to capture the relevant characteristics of a certain modality and convert them to a similar latent to the target domain. This conversion process is performed through a MS-SPADE (Multiple Switchblade Spatially Adaptative Normalization), which adapts the latents of the origin domain to the style of the target domain latents, aiding the style transfer. This block applies a normalization to the latents that adjusts $\mu$ and $\sigma$ of the features depending on the target modality. Each modality has its own set of normalization parameters. Additionally, these transformations are adaptative and spatial. Instead of an uniform transformation of the latents, the block adjusts different areas according to what is needed to match the target style. The block applies modulation functions to each pixel of the origin latent that locally adjusts it so that it adopts the style characteristics of the target. This modulation is unique to each modality. The idea of this architecture is that the modules of the MS-SPADE block can be changed according to the target modality that we want to transform to, allowing transformation to several different modalities using one single model, instead of using one model for each translation. Comparison to other state-of-art techniques such as pix2pix or CycleGAN show a better metric performance in T1-T2 and T2-FLAIR translation.

Other works aim to use image-to-image translation in order to improve the quality and resolution of MR images. Authors in \cite{wang2023spatial} claim to demonstrate that parameterization of conditional GANs in terms of spatial-intensity transformations (SIT) improves image fidelity and robustness to artifacts in medical image-to-image translation. Their SIT model can be seen as a way to manipulate latent representations through spatial and intensity transformations. The input images are codified into a latent representation using a generator. The spatial deformations applied to this latent code allows the model to perform morphological changes in the image, which is understood as a more precise representation of anatomical variations, meanwhile the intensity transformations allow the model to adjust the features of intensity. By separating spatial and intensity transformations, the SIT model allows for a clear interpretation of how variations in latent space are translated into anatomical changes and intensity of generated images. This model achieves a significative improvement of quality of images, reducing artifacts and improving anatomical fidelity. They evaluate the model using a longitudinal prediction of neurodegenerative patients task, where the model was able to predict trajectories of brain aging. Moreover, the model showed robustness in multi-center clinical data. The SIT framework provide a decomposition of anatomical and textural changes in the brain through latent representations that promise to be useful in a clinical setting.

Authors in \cite{zuo2021unsupervised} propose a model to perform harmonization between sites, named CALAMITY (Contrast Anatomy Learning and Analysis for MR Intensity Translation and Integration). This methods tackles contrast variability across multi-site MR images. The model is based on the information bottleneck (IB) theory and it employs a codification network to learn a disentangled latent space that separates anatomical and contrast information. The model is composed of an encoder (one for anatomy $E_{\beta}$ and another for contrast $E_{\theta}$, a decoder, and a $D_{\beta}$ discriminator, which distinguishes anatomical features according to the site, which helps to keep a global and uniform representation of the anatomy for all sites. The main idea is that the anatomical latent representation $\beta$ should only contain anatomical information, while the $\theta$ representation only contains contrast information specific of each site. To force $\theta$ to only learn contrast information, they train using two images of the same subject at a certain site but of different modalities. The only useful information is the common contrast, because both images represent the same anatomy, and so they expect $\theta$ to learn contrast information. In addition, if the CALAMITY model is trained on sites A and B, it can be fine-tuned to be applied in site C by using a small subset of its images, which allows for practical generalization. They found that their model outperforms baseline (no harmonization) and other harmonization techniques in the literature. However, the authors point that a limitation of the model is that the model assumes that T1-w and T2-w of the same subject capture the same anatomy, which is not always true. 

Building on this approach, authors in \cite{cackowski2023imunity} propose ImUnity, a VAE-GAN-based harmonization model that introduces several novel elements. Unlike CALAMITy, ImUnity takes two 2D images of the same subject from different locations of the brain and feed them as input. The first image is used to represent the anatomical structure, while the second one provides contrast information. The latent representations of each, obtained with a VAE, are concatenated to a single latent space, while the discriminator tries to differentiate between harmonized and real images demanding the VAE to produce better representations. On the other hand, the confusion module is connected to the latent space and tries to predict the original site of the image. A confusion loss is used to unlearn domain information, forcing the encoder to learn a representation that is invariant across domains. The biological preservation module is not mandatory and is used to introduce additional features such as sex or presence of diseases. Next, in order to create a 3D model, they use a 2.5D approach, where the model is trained in the three perpendicular axis and the results are combined along each axis. To validate the model, they perform three different experiments. In the first experiment they evaluate the harmonization task using traveling subjects from OASIS and SRPBS datasets. They compare the harmonized images and the real ones and use different performance metrics such as SSIM and histograms, showing that  ImUnity significantly improves the SSIM value compared to a CycleGAN and CALAMITY \cite{zuo2021unsupervised}. During the second experiment they classify the site of the MRI, where they found that the results of the harmonized data was superior to the non-harmonized ones, diminishing the capacity fo classify the origin of the site, suggesting the elimination of noise. Finally, for the third experiment they evaluated whether the model is able to improve classification of ASD (Autism Spectrum Disorder). They observed that in most cases, the AUC (Area Under the Curve) improved when performing cross-validation in several sites, from 2 different sites to 11, evaluation classification before and after harmonization. They found that in every combination of sites the AUC improved after harmonization.

The authors in \cite{wu2024disentangled} further advance harmonization by addressing machine-induced noise. In this work, they claim that MRI suffers from a huge machine-induced noise. They discuss that most errors and noise in MRI do not come from biological problems but from differences in scanners across different institutions and studies. To address this issue, they propose a model to harmonize MR images and eliminate the inter-site artifacts. The goal is to reduce the non-biological variations by using a style translation. Their DLEST (Disentangled Latent Energy-Based Style Translation) consists of three modules: 1) site-invariant image generation (SIG), site-specific style translation (SST) and 3) site-specific MRI synthesis. The first module uses an AE to codify the MRI into a latent representation, such that the anatomical content can be reconstructed with specific information from the site. This module aims to keep anatomical information. Then, SST applies a energy-based model to the latent space in order to modify its style from the origin site to the style of the target site. The last module generates synthetic MRI data with the desired style, useful for augmentation techniques. The main idea of the energy-based model is to learn the distribution of the target site and change the latent representations from the origin site to the target distribution. To do that, the model assigns values of energy to different configurations of the latent variables using $P_{\theta}(z_y)\propto e^{-E_{\theta}(z_y)}$, where $E_{\theta}(z_y)$ is the energy function that evaluates how compatible is a latent representation to the style target distribution. The goal is that the configurations closer to the target distribution have lower values of energy. Optimization is performed using Stochastic Gradient Langevin Dynamics, which simulates a random walk to low energy regions. To validate their results they use the public databases OpenBHB abd SRPBS. They compare the DLEST model to several models: non-learning techniques (Histogram matching HM and Spectrum Swapping), GAN models (CycleGAN) and ImUnity \cite{cackowski2023imunity}. They perform four different evaluation tasks to compare the performance of their model to the rest. The first task consists of histogram comparison and visualization of features, where the DLEST model shows a better alignment of histograms, specially in the white matter, the gray matter and CSF intensity pikes. In the second task they evaluate if the model is able to eliminate enough information related to the site by using a classifier. A clasfying network is trained to distinguish images from different sites. The DLEST achieves a balanced accuracy of $0.336$, significantly lower compared the other methods, which implies that the model is able to erase most of the irrelevant information. Next, a segmentation task shows that the model obtains consistent and robust results during segmentation, outperforming traditional methods. Finally, they evaluate whether the model can generate synthetic MRI data that is able to mach the style of a certain site. DLEST obtains high values of SSIM and Pearson correlation coefficient, similar to the ones obtained using pairs of real images from the site, meaning that the model captures the site variability correctly.

Using the same a anatomical/contrast latent space framework, authors in \cite{dewey2020disentangled} propose a harmonization model that leverages the inversion of the MR signal equation. This signal equation describes how MR images are generated from the interaction of the tissues and the magnetic field. Inverting this equation implies decomposition of the MR image in its fundamental parts \cite{jog2015mr}. Authors in this work use this inversion to create a disentangled latent space composed of two components: an anatomical latent space $\beta$ and a contrast representation $\theta$. To do so, they use a encoder-decoder model based on U-Net, where the decoder takes the concatenation of $\beta$ and $\theta$ to produce synthetic images. To encode the anatomical representation they use one-hot encoding $H\times W\times C_{ \beta}$, with $H$ and $W$ being the height and width of the image; and the one-hot encoding being along a channel $C_{\beta}$. This encoding allows restriction of information that can be passed through this latent space, since a flexible $\beta$ may pass extra information about the contrast. On the other hand, contrast is encoded into a vector $C_{\theta}$. To train this model they network is fed with a pair of T1-w and T2-w images, which contain the same anatomy, and so their $\beta$ representations must be equal, while the contrast maps will be different, due to the decomposition of the MR signal equation. In order to validate their model they trained it using 90 subjects from three different sites. Using SSIM and PSNR they computed the performance of the model by comparing harmonized images with real scans from traveling subjects. Using these subjects, they compared the scanned image in site A and its harmonized counterpart with the scan in site Be, finding that the harmonized images have a dramatic increase in the SSIM compared to the real image.

In \cite{qu2019wavelet}, the authors propose a model for translating 3T MRI images to 7T MRI images using adversarial learning. They highlight that a common issue with supervised methods is their reliance on a large number of paired 3T-7T images, which are difficult to obtain. To address this, they introduce a wavelet-based semi-supervised adversarial network that leverages the wavelet transform, which decomposes images into various frequency bands and spatial locations. This approach enables the analysis of the image structure at multiple levels, capturing both fine-grained details and global features. The GAN operates in this frequency domain, incorporating both spatial and frequency information, allowing the model to synthesize details at various frequency levels while preserving the high-resolution anatomical features typical of 7T images. The total loss function includes an adversarial loss that matches the distribution of synthetic images with that of the target images, a pair-wise loss that uses 3T-7T pairs to measure pixel-wise differences between the generated and real 7T images, a cycle consistency loss that ensures the transformation from 3T to 7T and back to 3T results in a matching image, and a wavelet loss that enforces the preservation of structural details.

To validate their model, the authors use 15 pairs of 3T-7T images and assess performance using the PSNR and SSIM metrics. They compare their results to those of four other 7T image synthesis methods: MCCA \cite{bahrami2016reconstruction}, RF \cite{bahrami20177t}, CAAF \cite{navab2015medical}, and DDCR \cite{zhang2018dual}. Their model outperforms the others in both metrics. Additionally, they conduct an ablation study to evaluate the contribution of each module to the model's performance. They demonstrate that omitting the pair-wise loss significantly reduces performance, while removing the cycle consistency loss leads to a decrease in PSNR values. Finally, they conclude that the wavelet loss is essential for capturing fine anatomical details.

\subsection{Visual reconstruction using fMRI}

Image reconstruction is a topic that has gathered significant attention in recent years. In computer vision we receive an image (e.g. a natural image) and try to either reconstruct that image (for image synthesis) by learning its distribution, or identify different elements within the image (for segmentation or parcelling tasks). In neuroimaging, researchers haven been trying to unravel the neural encoding of visual information by characterizing how this information $X$ is codified by the brain into latent representations $Z$. In general, the idea is to find the distribution $p(X|Z)$ that leads from visual input to neural representations that are later used to generate cognitive processes and responses. To do so, researchers use the information from fMRI images to see how different parts of the brain react to visual input. These activation maps contain relevant information of the latent representations, which can be analysed to understand the encoding process.

For instance, authors in \cite{st2018generative} use a GAN approach model to reconstruct natural images from fMRI. Here, they use a codifying model based on a feature-weighted receptive field (fwRF) model, described in previous work \cite{st2018feature}. This model predicts brain activity in response to visual stimuli, generating an activity vector $V$ from an image $X$, allowing prediction of fMRI responses to new images. An AE is applied to reduce dimensionality of $V$, generating a latent code $C$ that captures the relevant features of the brain activity. A conditional GAN is trained using the latent representation. The GAN is trained to generate images that are consistent to such code $C$.

They found that the model accurately predicts brain activity in response to natural images, and recovers well-known patterns of the brain in visual processing. They also found that reconstructed images were not easily recognizable by eye, but relevant features such as dominant lines were preserved by the model. 

Authors in \cite{seeliger2018generative} propose a method to reconstruct natural images from brain activity measured through fMRI scans using a GAN-based approach. During the experiments, subjects are presented with natural images from a dataset while their brain activity is recorded. A GAN is first trained on the dataset of natural images to learn a latent space capable of generating realistic reconstructions. Simultaneously, a predictive model is trained to map the recorded fMRI signals to the GAN's latent space. To achieve this, fMRI data are used as input to the predictive model, while the corresponding latent space coordinates of the GAN serve as the target labels.  

This setup allows the predictive model to learn the relationship between brain responses and the latent space coordinates of the GAN. Once trained, the system is tested by presenting subjects with novel images not included in the original dataset. Their fMRI responses are processed by the predictive model to generate latent space coordinates, which are then fed to the GAN. The GAN subsequently reconstructs an approximation of the presented image based on the derived latent space representation. The main idea is that the fMRI responses are linearly related to the latent representations. To validate their model, authors use three different datasets: BRAINS, that contains hand-written numbers, vim-1 with gray-scale natural images, and Generic Object Decoding, with gray-scale images of objects. They found that general structured of the images was reconstructed by the model but the quality depended on the dataset. For BRAINS dataset the reconstructed images were recognised by the subjects in $54\%$ of the cases, while in vim-1 and Generic Object Decoding was a $66.4\%$ and $66.2\%$ respectively.

Authors in \cite{han2019variational} propose a very similar model but with a VAE approach to reconstruct natural images from fMRI responses from subjects passively watching natural videos. The VAE is used to extract latent representations of every video frame, while a regression model is trained to link the fMRI response to the latent space, like in the previous work. The training of the regression model is simple. Since during training we know which video is the subject watching, we can link the fMRI obtained with the latent space of the video generated by the VAE. Once the model is trained we can show the subjects new videos, obtain their fMRI, use the regression model to estimate the latent space, and decode it to obtain the reconstructed video. Using fMRI to reconstruct the videos, they found that the reconstructions were blurry but were able to capture relevant features such as position, approximate shape of the objects, and contrast. To further validate their model they used the inverse process. Using the latent representation of a video, they used the regression model to predict the cortical activity. They found that the early visual areas (V1, V2 and V3) showed a higher prediction performance compared to the higher visual areas. However, the authors claim that the model demonstrate a modest level of success.

To address this issue, the authors in \cite{takagi2023high} propose using a latent diffusion model (LDM) approach to reconstruct natural images from brain activity. This method leverages not only visual information but also semantic representations to enhance the fidelity of the reconstructions. Using the LDM, they achieve high-resolution images with substantial semantic fidelity. The process begins with obtaining a latent representation $z$ using an autoencoder (AE) trained on natural images, while a linear model learns to map the fMRI data to the latent representations. The AE decoder then reconstructs an initial image $X_z$, which possesses approximate visual fidelity but lacks fine semantic details. Next, the authors introduce a textual representation $c$ linked to the fMRI responses from high visual areas. This representation encodes semantic information (in contrast to the structural details typically found in lower visual areas). The model subsequently combines $z$ and $c$ as inputs in a diffusion process based on a U-Net architecture. Finally, the AE decoder generates the reconstructed image $X_{zc}$, which incorporates both the visual information from lower areas and the semantic fidelity contributed by high visual areas. 
To validate their model, the authors conducted an ablation analysis using three configurations: a model with only $z$, one with only $c$, and a model integrating both $z$ and $c$. As expected, the model using only $z$ provided a good general visual structure but lacked semantic fidelity. Conversely, using only $c$ resulted in images with rich semantic content but poor visual precision. When both $z$ and $c$ were used, the model successfully captured both visual and semantic content. For better interpretability, the authors examined the correspondence between the representations and brain activation patterns using L2-regularized linear regression. They found that $z$ representations correlated with patterns in early visual areas (associated with borders and textures), while $c$ representations were linked to late visual areas (associated with semantic information). The combined $zc$ representations exhibited activation patterns balanced across both early and late areas. Additionally, to enhance interpretability, the authors analyzed the progression of $zc$ representations throughout the diffusion process. They observed that visual information is reconstructed during the early stages, while high-order semantic details emerge in the later stages. 

As the presence of semantic information has proven to be key in reconstructing natural image from fMRI, the authors in \cite{gu2022decoding} further improve this by proposing Cortex2Image, a framework that aims to reconstruct natural images from fMRI data with high semantic fidelity and detailed features. The model comprises two main components: Cortex2Semantic and Cortex2Detail, integrated with a pre-trained IC-GAN generator. Cortex2Semantic employs a spherical convolutional network to map fMRI brain responses to semantic image features, leveraging spherical kernels to account for the spatial topology of the cortical surface. The fMRI data is represented as multichannel data on a standardized icosahedral mesh of the brain, enabling feature extraction related to semantic categories using residual blocks and dense layers.

Cortex2Detail complements this by mapping brain responses to fine-grained image details, such as color, size, and orientation, using a variational approach to model the inherent uncertainty in fMRI data. The combination of these components with IC-GAN facilitates the generation of high-fidelity images by integrating semantic and detailed features. Importantly, training Cortex2Detail alongside the IC-GAN generator in an end-to-end manner significantly reduces computational overhead compared to previous optimization-heavy methods. 

To validate the model, the authors compared Cortex2Image with RidgeSemantic (linear ridge regression), RidgeImage (linear regression that predicts both semantic and fine-grained details) and Cortex2Semantic. They used both high-level (latent space distance of SWAV and EfficientNet-B1) and low-level (SSIM and pixel-wise correlation) metrics. They found that the model achieves high semantic reconstruction, such as zebras on grass or a train on tracks. However, fine-grained details are harder to achieves, but the model is able to reconstruct some of them, like the black strips of the zebras, the orientation of a plane. They also found that most faileres were due to the presence of multiple objects or categories within an image, or due to objects being partially occluded. Also, they performed an ablation study comparing the full model to Cortex2Semantic, finding that the later model is not able to capture fine details like orientation or spatial distribution of the objects.

The authors in \cite{ozcelik2023natural} introduce a model that outperforms the two previous works \cite{takagi2023high} and \cite{gu2022decoding}, which consists of a VDVAE and LDM approach named Brain-Diffuser to improve both semantic and visual reconstruction of fMRI reconstruction of natural images. VDVAE is a hierarchical VAE model that introduces several dependent latent variable layers. This VDVAE is used to extract low-level features from the images, which serve as an initial guess. Next, as usual, they use a regression model to link the fMRI responses to the latent space. In the second stage they use a latent diffusion process conditioned on predicted multimodal (text and images) features using CLIP-V and CLIP-T \cite{radford2021clip} which link the visual and textual representations into a common space. Finally, the reconstructed image contains both visual and semantic information.
In general, the authors found that the reconstructed images  capture most of the layout and semantics of the ground truth. However, like the previous work, the Brain-Diffuser approach fails to reconstruct under certain situations such as occlusion or objects. They compare their results to \cite{takagi2023high}, \cite{gu2022decoding}, using SSIM, AlexNet(2), PixCorr, InceptionV3, CLIP, EffNet-B and SWAV as performance metrics. Authors claim that theirs is the best performing model by a decent margin for all the quantitative metrics. Next, an ablation study showed that VDVAE have a better performance but only for the low-level metrics while it obtained the worst results for the high-level metrics. On the other hand, the Brain-Diffuser model without the VDVAE is not able to capture visual structure, suggesting that the inicial guess of the VDVAE is necessary but not sufficient. Another ablation study showed that CLIP-T aids to achieve both low-level and high-level performance. The high-level improvement of CLIP-T is expected since text is used to add semantic fidelity, however, the low-level improvement can be explained due to the layout of the image (such as number of objects and orientation, which can be added to the text). Next, ablation of CLIP-V does not reduce low-level information as much as expected. Finally, to improve interpretability of the model, the authors perform a ROI analysis to see the relation between brain region and the components of the model. The found that V1-V4 visual regions carries more information from the VDVAE features, while specific regions (words, faces, places, etc.) carries more information from the features of CLIP.

 Authors in \cite{guo2024mindldm} propose a model with high semantic and visual information, outperforming all the previous state-of-the-art works that we have previously discussed. The MindLDM model is based on LDM and VAE, used for reconstructing natural images from fMRI using both visual and semantic information. The model has 4 stages: pretraining of fMRI feature extration, alignment of fMRI and text in CLIP space, generation of depth maps using VDVAE, and LDM. First, latent representations of the fMRI data is obtained using a MAE encoder. Next, using CLIP common feature space, multi-tag text is aligned to fMRI embeddings. In order to obtain better visual reconstructions, a VDVAE is trained to obtain depth maps from the natural images and, using a regressor, depth maps can be obtained from the fMRI input images, which are used to guide the generation of the images. Finally, a diffusion process generates the image conditioned on the fMRI latent, the text and the depth map.

 The authors validate the model using both low-level and high-level performance metrics. For the low-level metrics, they evaluate the similarity of basic features between reconstructed image and ground truth, using SSIM, PixCorr, AlexNet(2) and AlexNet(5), which compare features such as borders or texture. For the high-level evaluation they use InceptionV3, CLIP, EffNet-B and SWAV. They compare their model to \cite{gu2022decoding}, \cite{takagi2023high}, \cite{ozcelik2023natural}. Quantitative results show that MindLDM outperforms the other models in key metrics, showing higher values in high-level metrics and good results in low-level metrics, suggesting high precision in both structural and semantic similarity. Furthermore, the authors also perform an ablation study by removing elements such as the fMRI/text alignment or the use of depth maps to guide reconstruction. They found that the multi-tag approach was more effective than natural language. Also, the study showed that the fMRI/text alignment is crucial to the model, leading to large semantic deviations if not implemented. On the other hand, depth maps proved to be necessary to achieve structural reconstruction. This ablation study indicates that every part of the model is necessary to reconstruct the images.

 \subsection{Brain ageing analysis}

 The brain undergoes a complex and multifaceted process shaped by changes in both its structure and cognitive function throughout an individual’s lifespan. This process is further influenced by factors such as sex, age, and external elements like social interactions, making the modeling of brain changes over time a challenging task. Despite these challenges, such modeling is essential for distinguishing between normal brain evolution and the onset of neuropathologies. For example, analyzing aging patterns can help determine whether an older individual is experiencing typical age-related cognitive decline or the early stages of a neurodegenerative disease. Similarly, it can provide insights into the developmental stages of the neonatal brain, enabling the identification of potential issues in early brain development. 

 Authors in \cite{xia2019consistent} use a VGG-like network consisting of convolutional layers with small kernels to synthesize 2D MRI images of subjects at older age $t$, given a ground truth MRI $t_0$. They represent age as ordinal vectors, ensuring that difference between vectors is correlated to difference between ages. The model consists of a generator based on a transformer, which generates the samples at $t$ from an input at $t_0$, and a discriminator that is trained to distinguish the ground truth at $t$ and the generated image at the same age to compare the performance of the model. The generator takes the image $x_{t_0}$ as input encodes it to obtain a latent representation, concatenates with the age difference vector $v_d$, apply the transformer, and outputs the image $\hat{x_{t}}$.  This image is fed into the discriminator along with $v_d$ and evaluates whether the image looks real and consistent with the target age. The total loss comprises the adversarial loss, $L_{GAN}$, and an identity preservation loss, $L_{ID}$. The latter ensures that the identity of subjects remains intact throughout the aging process by assessing whether the difference between $\hat{x}_t$ and $x_{t_0}$ is within acceptable bounds for a given age difference $d$.
 
Due to the insufficient availability of longitudinal data, the authors train a predictor, denoted as $f_{pred}$, to estimate the age of the subjects in the dataset. The performance of this predictor is evaluated by computing the difference between the predicted and actual ages. The results indicate that brain changes are minimal during the early stages of aging. However, after the age of 52, significant brain changes occur, capturing features that align with existing literature on brain aging. 

In addition, the authors conduct a longitudinal analysis using a small subset of the ADNI dataset. They report a mean absolute error (MAE) of 0.08 for their model, which outperforms a conditional GAN-based approach with an MAE of 0.21, and a CycleGAN approach with an MAE of 0.20. These results suggest that the images generated by the proposed model are realistic and effectively preserve the identity of the subjects.

Authors in \cite{choi2018predicting} use a VAE approach using PET scans and age data of the subjects. In this work, they hypothesize that latent features of the PET scans do not change with time. To VAE take the input PET at time $t_0$ and obtain its latent representation, which is then concatenated to the future age $t$ and decoded to obtain the reconstructed image $\hat{x}_t$. To validate the model they compared the follow-up PET of the subject to the reconstructed image using delta maps. They found that the average predicted changes of the brain were correlated to the real changes of the follow up. The delta maps were positively correlated with the delta maps of the real images. This seems to indicate that the model was able to capture change patterns of brain metabolism that align to the observed real changes. Moreover, the study showed a good capability to capture subject variability. Some subjects showed a global diminishing of metabolism suggesting normal ageing or cognitive deterioration, while other subjects showed an increase of metabolism in specific areas such as the frontotemporal cortex, which may indicate a compensation of the metabolism. In general, the latent representation were able to provide information about how the brain metabolism changes over time. 

In addition, the authors developed a model that added APOE4 state information along with the age to the latent representation in order to see how it affects the development. They found that adding APOE4 produced a lower metabolism in the hippocampus and amygdala. From the age of 60 onwards they found that metabolic differences were less pronounced, suggesting that changes related to APOE4 are more significative in the early stages. They also found using delta maps that APOE4 related chances showed a rapid decrease in metabolism in the occipital areas.

Authors in \cite{chadebec2022image} propose a longitudinal model for generating synthetic images that simulate patient trajectories, filling in the missing data between patient visits, extracting features of MRI images using a VAE model. The trajectory of a patient is modeled through a linear equation 
\begin{equation}
    l_i(t) = e^{\eta_i}(t-\tau_i)\cdot e_1 + \sum_{k=2}^d \lambda_i^k\cdot e_k,
\end{equation}
where $\eta_i$ is a velocity parameter that quantifies the progression of the individual, $\tau_i$ is the delay, and $\lambda_i$ are spatial parameters that determine the shape of the trajectory in the euclidean space. The model uses a variational approach where the latent variables represent the parameters of the trajectory equation, learning the distribution of such variables. They assume that those parameters area independent for each subject. If we image modeling of an AD subject, $\eta_i$ would indicate velocity of the progression, $\tau_i$ the onset of the degeneration, and $\lambda_i$ how the changes affect the space (e.g, size of the lesion). The authors assume that the latent space is factorized as $q_{\phi}(z_i|x_i) = q_{\phi}(\eta_i|x_i)q_{\phi}(\tau_i|x_i)q_{\phi}(\lambda_i|x_i)$. Once the trajectory $l_i(t_j)$ has been evaluated, the decoder of the VAE transform the features into a reconstructed image.

The authors validate the model using MSE and SSIM performance metrics. Using the ADNI dataset, they found that the trajectories showed an increase of the ventricles over time, which is an indicator of the progression of the disease. The authors claim that although the trajectories seem to be realistic, the model requires more validation by expert physicians.

Authors in \cite{calm2024identifying} use a VAE coupled with a regression model to identify age-related patterns. The regression model uses input data to predict the age. The loss term includes the usual ELBO loss of the VAE and an extra term for the regression that measures how precise is the model predicting age, by using the KL divergence between the computed age distribution $q(c|x)$ and the real age distribution $p(c)$. In the latent prior they include a term $p(z|c)$ to assure that the latent representations $z$ are influenced by the age $c$.

The authors train two different models, one for female subjects and a second model for male subjects. In order to interpretate the latent space they use SAGE (Shapley Additive Global importance) fro each dimension, that quantifies the predictive power of each independent feature of the latent space and compute the contribution of each region to the prediction. Also, they used OLS (Ordinaty Least Squares) regression to estimate the relation between each risk factor and each latent variable. They found that the VAE+regression model outperforms VAE+MLP model. Using SAGE, they where able to identify the characteristic regions vulnerable to dementia, which where then linked to risk factors and trajectories related to age, clinical and cognitive variables. Using the OLS, they found differences between the female and the male model, where males had more correlations indicating  good health, whereas the female model showed less correlations but an increase of the risk factors. 

In addition, they found that both their VAE-regression model and the VAE-MLP overestimated age of young individuals and underestimated age of old subjects. Moreover, the were able to identify four different key dimensions in the female model, some of them related to dementia (frontal and parietal lobes, and hippocampus). In the male model they also found four dimensions, some related to dementia (temporal (entorhinal) and parietal lobes, hippocampus and frontal lobe. In conclusion, the model is able to identify possible brain-related trajectories and associate them with dementia-related risk factors.

Authors in \cite{bieder2024modeling} use an Implicit Neural Representation (INR) approach to model and represent continuous functions like 3D images instead of representing data as pixels, to model the development of neonatal brain during the third semester of pregnancy. They model images as functions that depend on a space domain $\Omega$ (pixel coordinates), a normalized time $T$ and a latent space that represents the identity of the subject $\Lambda$, which define a function $f_{\theta}(x,t,z)$. During training, the intensity value of the image is computed using $f_{\theta}$ and minimized such that the MSE between predicted intensity $\hat{I_x}$ and $I_x$ are similar. The idea is that the latent vectors are the same for all the images of a single subject (SSL, Subject Specific Latent), like the previous subject, aiding the model to capture its identity, while time is used to model the development and ageing. In order to tackle the problem of limited longitudinal images for some subjects the authors use a Stochastic Global Latent Augmentation (SGLA) where a global latent is used during training for these subjects. This allows the network to learn to predict images at different ages. During training, optimization allows the disentanglement of identity and age. The global latent is shared among every subject with only one image and in every iteration there is a probability that the model will use the global latent instead of the specific subject latent. With this training approach the model is forced to learn how the features of the subjects can be generalized from the age $t$, allowing to simulate how the images would be during the different stages of the development even when there is no enough data.

To validate the model they compare to a denoising diffusion model with gradient guidance (DDM+GG). For each subject in the dHCP (T2 MRI) dataset they consider two scans taken at different temporal points. Using the PSNR, SSIM and MAE metrics they found that their model outperformed the diffusion model. They also performed an ablation study comparing the performance of the model with and without every combination of SSL and SGLA, finding that the best performing model was the one that included both methods. Using measures of the head circumference (HC) they found that the needed both SSL and SGLA to find correlation between HC of the predicted image and the ground truth. Qualitatively they found that the reconstructed images matched the size correctly but had some difficulties capturing details such as the exact shape of the folds, due to the bottleneck of the latent space.

\subsection{Disease classification}

Generative latent models, particularly Variational Autoencoders (VAEs) and their variants, have become pivotal in the field of neuroimaging for disease classification. These models are uniquely suited to extract structured, low-dimensional latent representations from high-dimensional neuroimaging data, capturing complex patterns while preserving critical information about disease-related variations.

The latent representations obtained from generative models offer several advantages: they provide a compact and interpretable feature space for classification tasks, enable the exploration of disease progression trajectories, and allow the integration of uncertainty into predictions. This section highlights recent advancements in applying generative latent models to neuroimaging for classifying neurological and psychiatric diseases, focusing on the extraction and utilization of latent spaces to enhance both predictive performance and interpretability. By leveraging these latent spaces, researchers not only improve classification metrics but also gain insights into the neurobiological underpinnings of disorders, paving the way for more personalized and precise diagnostics.

Authors in \cite{basu2019early} propose a VAE+MLP model to predict the status (Normal or Alzheimer) of a patient 6 months after a visit, using MRI. The VAE extract relevant features of the MRI which they use to see which brain areas contribute the most to the model and quantify the distribution of possible trajectories of the subject. To predict the status label they use a generative classifier that models the joint distribution of features and labels $p(X,Y)$.

The latent representations of the VAE are fed to the MLP to predict the status, using a cross-entropy loss along with the usual ELBO of the VAE. Using the VAE approach the authors obtain an accuracy of $74.40\pm0.01$ and a F1-score of $0.66$. Additionally, they perform a risk analysis. For each MRI in the test set, they take 100 samples from the latent space and predict the future disease status. They find that $59\%$ of the samples are not at risk, while the rest show varying degrees of risk. In contrast, the other models used for comparison (CNN and CNN-AE) predict that only $2.79\%$ and $9.16\%$ of the samples are not at risk, suggesting that the VAE approach provides a more robust prediction of risk compared to the other models. Qualitatively, using relevance maps, the authors found that the model focuses on anatomical specific areas for the prediction, such as the cerebellum, the neocortex and the brainstem, which are associated with the progression of Alzheimer's disease. They claim that their study predicts future progression, while other studies focus on correlating changes with current symptoms. However, they note that there are limitations in terms of which specific features the model is using within this regions, suggesting that further research is needed to understand the model's behaviour.

On the other hand, authors in \cite{zhou2019deep} use a multi-modal fusion of different data modalities, such as MRI or PET, using ADNI scans, to classify dementia patients. They aim to create a model that is both useful in clinical setting and interpretable. To perform the multi-modal learning the authors use a Negative Matrix Factorization (NMF) which is a technique that decomposes a data matrix into simpler components, with the property of each matrix being non-negative. If we have a matrix $X$ of dimensions $d\times n$ where $n$ is the number of samples and $d$ the number of features, the method targets to decompose $X$ into two matrices $B$, containing the fundamental features and $H$ that indicates how to reconstruct the data. The idea is that the product $B\cdot H$ resembles $X$. In the multi-modal case we have $B(v)$ and $H(v)$, one pair of matrices for each independent modality. However, using the model 
\begin{equation}
    min \sum_{v=1}^V ||X(v) - B(v)H(v)||^2,
\end{equation}
there would be no relation between different modalities. In order to address this issue, the authors use a common $H$ matrix for all modalities. In order to obtain deep latent representations, they use the deep NMF model, which consists on decomposing each matrix into lower representations:
\begin{equation}
    X^{(v)} \approx B_1^{(v)}\dots B_L^{(v)} H_L.
\end{equation}
Using these latent representations the authors classify Alzheimer patients in the ADNI dataset into three categories: Normal Control (CN), Mild Cognitive Impairment (MCI) and AD, utilizing MRI and PET scans. They compare their model to other state-of-the-art classification models, using accuracy, sensitivity, specifity and F-score as performance metrics. The found that the proposed model outperforms other methods such as SVM, MKL, shallow NMF, etc. in all metric values on all classification taks. They also found that all methods using multi-modality data outperforms all models using a single modality, demonstrating its robustness. In addition, they perform a  pMCI vs sMCI classification task, showing that the model shows the highest accuracy values compared to other state-of-the-art methods.

The authors of \cite{matsubara2019deep} propose a model designed to improve the diagnosis of psychiatric disorders, particularly schizophrenia and bipolar disorder, by addressing challenges such as overfitting and the extraction of irrelevant patterns common in classification tasks. The proposed model employs a deep generative model (DGM) that calculates the posterior probability of a subject's state based on their fMRI data. To extract features from the images, the model incorporates a modified VAE that integrates the diagnosis label (control or disease) directly into its structure. This integration enables the model to distinguish relevant patterns associated with the diagnosis from irrelevant variations, such as noise or frame-wise variability. The inclusion of the label as an independent variable explicitly conditions the model to associate meaningful features with the label while disentangling and ignoring irrelevant factors.

The authors compare their model against several baselines, including functional connectivity (FC)-based methods, autoencoders (AE), multilayer perceptrons (MLP), Gaussian Mixture Models (GMM), and dynamical models based on Markov chains and LSTM. The proposed model significantly outperforms the baselines in multiple performance metrics for schizophrenia, achieving a balanced accuracy of $85.3\%$, along with superior Matthews correlation coefficient and F1-score. Additionally, it identifies key brain regions, such as the thalamus, which is strongly associated with schizophrenia. For bipolar disorder, the balanced accuracy reaches $81.5\%$, with other metrics comparable to the baseline models. In this case, the regions most influential for classification are the cerebellum and several frontal regions, consistent with findings in the literature.

Authors in \cite{brodersen2014dissecting} divide psychiatric spectrum disorders into neurophysiologically defined subgroups using a generative embedding approach. They hypothesize that, within a Dynamic Causal Modelling (DCM) framework, patients with schizophrenia can be classified into subgroups corresponding to distinct clinical subtypes, particularly based on the severity of negative symptoms assessed by the PANSS-NS scale.

First, they extract temporal series from the visual, parietal, and prefrontal regions during a working memory task. The DCM framework is employed to infer effective neural connectivity across these regions, modeling the brain as a dynamical system where neuronal populations interact via synaptic connections. This generates a feature space that encodes effective connectivity parameters. Using Bayesian inference, the posterior distribution of these parameters is estimated for each subject, and these estimates are then converted into feature vectors representing the strength of specific connections. Each subject is embedded in a 12-dimensional space that encapsulates the architecture of their brain network under task conditions.

Using a support vector machine (SVM) classifier, they distinguish controls from patients with a balanced accuracy of $78\%$. Subsequently, applying Gaussian mixture models (GMM) for unsupervised clustering, they identify distinct subgroups. Restricting the analysis to patients reveals three neurophysiologically defined clusters with unique connectivity patterns. Critically, these clusters map onto differences in clinical symptomatology, specifically in the severity of negative symptoms, thus validating the relevance of their approach.

The authors in \cite{dong2020spatiotemporal} propose a Spatiotemporal Attention Autoencoder (STAAE) for ADHD classification by capturing long-distance dependencies of global features using fMRI data. The attention mechanism addresses the limitations of RNNs and CNNs when handling high-dimensional fMRI data. This mechanism facilitates the capture of global dependencies through the use of query, key, and value matrices \cite{vaswani2017attention}. The STAAE employs an encoder-decoder framework, where the encoder maps sequences of symbolic fMRI representations into intermediate representations. These representations are then utilized to estimate matrix coefficients via Lasso regression, constructing spatial maps that link latent features to brain regions associated with Rest State Networks (RSNs). By leveraging these RSNs, the authors create a brain atlas that combines the ROIs of all RSNs. Using this functional connectome derived from the RSNs, they classify ADHD subjects.

Visual inspection of these RSNs confirms their interpretability and alignment with the RSNs reported in ADHD literature. To benchmark their method against other baseline approaches, the authors use the overlap score, demonstrating that the STAAE accurately identifies intrinsic RSNs. Furthermore, they compare the classification accuracy of the STAAE with that of SDL, RAE, SVM, and ICA, showing that the STAAE outperforms these methods across all four datasets. However, the authors acknowledge that the effects of hyperparameters remain insufficiently explored, highlighting the need for further research in this area.

\subsection{Functional brain networks}

Functional brain networks (FBNs) play a critical role in understanding neural connectivity and activity. Several recent studies have leveraged deep learning techniques to uncover these networks from fMRI data. For instance, the work in \cite{qiang2020deep} introduces a deep variational autoencoder (DVAE) to derive latent representations from 4D fMRI data signals (3D volumes + temporal axis) associated with Functional Brain Networks (FBNs). Using the latent space, they apply Lasso regression between the original fMRI data (converted to a 2D matrix by concatenating time and voxel dimensions) and the latent variables. This generates a coefficient matrix that is mapped back to the 3D brain space. Each row of the matrix is transformed into a 3D image representing an FBN, effectively acting as a filter to identify the voxels where each latent variable is most relevant for functional activity. This approach links latent variables to specific FBNs.

Temporal series are then extracted from regions of interest (ROIs) defined by these FBNs, and Pearson's correlation coefficients between ROIs are computed to construct functional connectivity matrices for each subject. These matrices serve as input for supervised classifiers, including linear and radial SVMs, Random Forest, and a custom-designed deep neural network (DNN).

To validate the DVAE model, the authors compare it with Sparse Dictionary Learning (SDL) and a standard autoencoder (AE). The results show that DVAE outperforms AE in deriving meaningful FBNs, particularly in small datasets, as it avoids overfitting and captures more generalized features. Quantitative results demonstrate robustness across datasets and superior performance in three out of the five centers analyzed in the ADHD-200 dataset.

The study highlights the hierarchical organization of FBNs derived from different DVAE layers. Shallow layers capture simple connectivity patterns, while deeper layers reveal complex, global interactions. This hierarchy is validated through the Inheritance Similarity Rate (ISR), which measures how FBNs from shallower layers overlap with and contribute to those in deeper layers. Results indicate that multiple partially overlapping FBNs in shallow layers combine to form more complete FBNs in deeper layers, aligning with findings in the literature.

Finally, the DVAE pipeline demonstrates competitive performance for ADHD classification, achieving higher accuracy in some centers compared to other fMRI-based methods and offering robust results despite variability in dataset size and scanning parameters. The authors emphasize the DVAE's capacity to extract hierarchical and biologically meaningful features, making it a promising tool for fMRI data analysis.

The authors in \cite{zhang2020deep} propose the Deep Multimodal Brain Network (DMBN) model, which integrates structural and functional brain networks, representing them as graphs. The structural graph acts as a scaffold, imposing constraints on functional activity, while functional activity gradually influences the structural anatomy over time \cite{bullmore2012economy}. The DMBN model employs an encoder-decoder architecture to translate structural networks into functional representations. Initially, node representations are extracted using a convolutional kernel on the graphs, capturing connectivity patterns among neighboring nodes in the structural network, which are then used to learn the functional network representation. To enhance the model's ability to dynamically adapt connection weights, an attention mechanism is applied, enabling the identification of nuanced relationships between nodes. The model captures the non-linear and indirect relationship between structural and functional connectivity, learning an effective translation from one modality to the other. The decoder reconstructs the functional network from the structural node representation, capturing both direct connections and complex, non-linear interactions.

The authors benchmarked the DMBN model against five state-of-the-art methods, including three machine learning models (tBNE \cite{cao2017t}, MK-SVM \cite{dyrba2015multimodal}, mCCA+ICA \cite{sui2011discriminating}) and two deep learning models (BrainNetCNN \cite{kawahara2017brainnetcnn}, Brain-Cheby \cite{ktena2018metric}). DMBN outperformed all the baselines in a gender prediction task, achieving an accuracy of $81.9\%$ and a $10\%$ improvement in the F1-score. Using saliency maps, they identified the top 10 brain regions most relevant to the prediction task. These regions included cortical areas such as the orbital gyrus, precentral gyrus, and insular gyrus, as well as subcortical areas like the basal ganglia, which are critical for cognitive regulation, motor and emotional control, and likely exhibit gender-related differences. Moverover, an ablation analysis shows that the use of multimodal networks is essential to obtain a representative latent space, along with the attention mechanism.

Furthermore, the authors applied DMBN to a disease classification task using the PPMI database for Parkinson's Disease. The model demonstrated superior performance compared to baseline models, achieving a $5-9\%$ improvement in accuracy. Additionally, DMBN identified 10 key regions associated with Parkinson's biomarkers, such as the bilateral hippocampus and basal ganglia, which are well-documented in neuroimaging studies of Parkinson's Disease.

The study by Dong et al. \cite{dong2020discovering} employ a 3D residual autoencoder (ResAE) to model deep representations of fMRI data. Their approach simplifies functional brain network (FBN) estimation by using a deep autoencoder (AE) to learn latent representations of functional activity. These latent representations are subsequently used for FBN estimation via lasso regression, enabling the construction of a functional activity network matrix. Each row of the matrix maps to the 3D brain space, generating an anatomical spatial map. The authors compare the ground truth Hemodynamic Response Function (HRF), measured as the response to stimuli during tasks, with the temporal features generated by the ResAE, finding a high degree of precision in the match.

To further validate their approach, the authors compare the performance of their deep ResAE with a shallower ResAE, demonstrating that the deep ResAE extracts features that align more closely with the ground truth than those obtained by the shallower model.

Moreover, the temporal features extracted by the deep ResAE are mapped back to the MNI space to analyze the corresponding spatial maps. The study reveals that each hidden layer learns functional networks of varying complexity, depending on the depth of the layer. The authors find that the resulting FBNs are interpretable and consistent with findings in the existing literature. Using overlap rates, they show that the deep ResAE significantly outperforms the shallower model in capturing meaningful FBNs.

Finally, their model successfully identifies resting-state networks (RSNs) even during task performance. Notably, the model captures principal RSNs, including the visual, default mode, sensorimotor, auditory, executive control, and frontoparietal networks. This demonstrates that the model can dynamically identify both task-related FBNs and active RSNs.

\subsection{Multimodality integration}

Generative latent models have proven to be highly effective in integrating multimodal neuroimaging data by projecting diverse data sources into shared lower-dimensional latent spaces. These models enable the extraction of unified representations that capture complementary aspects of the underlying data while maintaining interpretability and scalability.

For instance, the Variational Autoencoder (VAE) proposed in \cite{geenjaar2021fusing} integrates functional and structural neuroimaging data into a shared latent space using a single encoder-decoder scheme. This approach facilitates the extraction of cross-modal patterns, allowing for interpolation and interpretability between modalities. By employing these shared latent representations, the model achieved superior performance in schizophrenia classification tasks compared to traditional fusion methods.

To evaluate the effectiveness of the representations, the authors conducted a schizophrenia classification task. The model achieved a receiver operating characteristic area under the curve (ROC-AUC) of $0.8609$, outperforming both eatly and late fusion PCA methods. The VAE's ability to reconstruct differences in the latent space was also validated through visualization of group differences between schizophrenia patients and healthy controls. Key regions, such as the thalamus and cerebellum, emerged as important, consistent with prior findings.

Moreover, the model demonstrated scalability with increasing latent dimensions, retaining meaningful information without overfitting, even in small datasets. Importantly, the latent space showed clusters corresponding to specific modalities, such as intrinsic functional networks (ICNs) and structural MRI (sMRI) supporting its robustness. The model's generative capability offers potential applications in data augmentation and further exploraation modality-specific interactions.

On the other hand, following the same multimodality approach, authors in \cite{martinez2024bridging} propose a joint VAE model that learns a shared latent representation of both 123I-ioflupane SPECT images and clinical data scores of Parkinson’s Disease (PD). Unlike previous work, they use two VAEs, one for neuroimaging and another for clinical data, with shared latent spaces cross-related by an additional loss term that minimizes differences between shared latents while ensuring disentanglement of non-shared latents. Moreover, the authors employ Maximum Mean Discrepancy (MDD) instead of the usual Kullback-Leibler divergence to maximize mutual information between input data and latent space.

Using this shared latent representation, the model achieved an $R^2$ of up to $0.86$ in same-modality tasks and $0.441$ in cross-modality tasks for predicting motor symptomatology and clinical features such as UPDRS. The framework demonstrates the feasibility of bridging neuroimaging and clinical modalities, identifying latent features predictive of motor symptoms and cognitive deficits. Notably, their analysis revealed that neuroimaging-based latent features are predominantly specific to motor impairments, reflecting dopaminergic deficits characteristic of PD. The study underscores the importance of data standardization and DSSIM for enhancing latent representations, making a significant contribution to the understanding of multimodal neurodegenerative patterns. Future directions include early-stage detection and extending the framework to other neuroimaging modalities.

In \cite{ghosal2019bridging}, the authors propose a model to integrate imaging and genetic data into a lower-dimensional manifold guided by clinical diagnosis. Each subject in the framework has three inputs: an fMRI activation map, genetic information, and a binary diagnosis (control or affected). The model employs two feature matrices, $A$ and $B$, representing imaging and genetic data, respectively, along with a shared latent vector $z_m$, unique for each subject but common across both modalities. This setup models the imaging data as $f_m \approx A^T z_m$ and the genetic data as $g_m \approx B^T z_m$. The matrices $A$ and B represent the anatomical and genetic basis shared across subjects, while $f_m$ and $g_m$ project the specific features of each subject into the latent space. Using the shared latent representation $z_m$, the authors predict the probability of a subject being affected by the condition via logistic regression, $\sigma(z_m^T c)$ where $c$ is a regression coefficient vector. The sigmoid function ensures probabilities near one for high $z_m^T c$ values. To validate their approach, the authors compare four models: RT (random forest), CCA+RT, Img (only the imaging subspace), and Img+gen (imaging and genetic data). Using sensitivity, specificity, and accuracy as metrics, the Img+gen model outperforms others. They identify key regions from the matrix $A$, including the dorsolateral prefrontal cortex, implicated in executive function deficits characteristic of schizophrenia, as well as the hippocampus and parahippocampus, regions commonly affected by the disorder. Additionally, the model highlights five single nucleotide polymorphisms (SNPs) most relevant to the latent space. The authors perform a genome-wide association study (GWAS), confirming that these SNPs are part of genes linked to schizophrenia. This demonstrates the model's ability to uncover clinically relevant biomarkers integrating imaging and genetic data.

In \cite{jeon2024gene}, the authors propose a more direct approach to integrating imaging and genetic data by leveraging an implicit generative model based on Latent Diffusion Models (LDM). Their framework generates MRI data conditioned on labels and genotypes. First, an autoencoder extracts latent representations from MRI slices. These representations are then fed into a diffusion process conditioned on both labels and genotypes. Labels are concatenated with the latent space, while genotypes are incorporated via a transformer. The use of cross-attention enables the model to focus selectively on relevant genetic features, guiding the diffusion process to create meaningful new representations.

The model was validated on the ADNI dataset and compared against five frameworks: CGAN, StackGAN, AttGAN, a label-conditioned LDM, and several variants of the proposed LDM (G2I, G2I-1, and G2I-1+Att). Evaluation metrics included PSNR, FID, SSIM, and MS-SSIM. Results showed that the full model (G2I-1+Att) outperformed all competitors, generating images closely resembling real data and accurately capturing key markers of Alzheimer’s progression, such as ventricular enlargement and cortical atrophy. Notably, only LDM-based models consistently produced complete and realistic MRI scans, whereas GAN-based models struggled with generating fine details. While AttGAN achieved the highest PSNR score, its outputs appeared blurry, highlighting that PSNR alone may not fully capture visual quality.

An ablation study further demonstrated that incorporating genetic data via cross-attention significantly improved image quality compared to the baseline, underscoring the importance of this design choice.

\subsection{Image synthesis}

The generation of synthetic medical images has emerged as a vital tool for addressing challenges such as data scarcity, privacy concerns, and the need for controlled experimentation. Among the most popular frameworks for image synthesis are Generative Adversarial Networks (GANs) and Latent Diffusion Models (LDMs), which have proven particularly effective in medical imaging and neuroimaging applications. These models are widely used to enhance image realism, diversity, and structural fidelity, playing a critical role in generating data for training algorithms and simulating complex anatomical variations.

Zhang et. al \cite{zhang2019skrgan} present the Sketching-rendering unconditional Generative Adversarial Network (SkrGAN), an innovative approach that incorporates a sketch-based structure to guide the image generation process. By decomposing the generation into a sketch module ($G_s$) for structural outlines and a color rendering module ($G_p$), the model preserves essential anatomical features, such as edges and boundaries, derived through Sobel edge detection. This framework demonstrates the adaptability of GANs in generating synthetic images that retain critical structural characteristics, highlighting their utility in medical imaging.

They use MS-SSIM, Sliced Wasserstein Distance (SWD) and Frechet Inception Distance (FID) as performance metrics, where the SWD measures the distance necessary to transform a probability distribution into another one. On the other hand, FID compares the statistical features (such as $\mu$ or $\sigma$) of both real and generated images. 

Expanding on the limitations of traditional VAEs and GANs, authors in \cite{pinaya2022brain},introduce a latent diffusion model tailored for generating  high-resolution 3D brain MRI scans. LDM combines the efficiency of autoencoder-based latent representations with the probabilistic modeling capabilities of diffusion models. The authors trained the model on 31,740 T1-weighted MRI scans from the UK Biobank, compressing input data to a latent space before applying the diffusion process. This approach enabled stable generation of high-quality images while managing computational demands.

Model performance was evaluated using metrics such as Frechet Inception Distance (FID), MS-SSIM, and 4-G-R-SSIM, where lower scores denote better diversity and realism. The LDM outperformed baseline methods like LSGAN \cite{mao2017least} and VAE-GAN \cite{larsen2016autoencoding}, producing sharper and more realistic images. Additionally, leveraging a hybrid conditional approach with cross-attention mechanisms, the model could generate images conditioned on attributes like age, sex, ventricular volume, and normalized brain volume. Ventricular volumes, assessed with SynthSeg, showed a Pearson correlation of 0.972 between generated and conditioned values, demonstrating strong alignment. For age prediction, a 3D convolutional neural network trained on the generated images achieved a correlation of $r=0.692$ with input values.

The authors also tested the model's ability to extrapolate conditioning variables beyond training ranges, producing plausible outputs, such as abnormally large or absent ventricles. To foster reproducibility and research, they released a dataset of 100,000 synthetic brain MRI scans via Academic Torrents \cite{pinaya2022dataset}, Figshare, and the HDRUK Gateway. This work highlights the potential of diffusion models using latent representations in overcoming data scarcity in medical imaging while preserving privacy.

\subsection{Other works}

Additionally to the ongoing lines of investigation mentioned in this review, some other authors have conducted various studies to further enhance the processing, analysis, and understanding of neuroimaging. These works vary in topics and include studies about generation of 3D MR images from 2D slices, increasing of resolution of MRI, interpretability of latent representations, the use of genetic information along with imaging, or even the use of EEG signals in the context of latent representations.

For instance, authors in \cite{volokitin2020modelling} use a 2D VAE to learn representations of 2D brain MRI, and a Gamma model that captures relationships between slices to generate 3D MRI. By estimating the mean and covariance in the latent space of the 2D model in the direction of the slice they can obtain the distribution along the slice direction and generate the remaining slices. They also use a validation method for volumes that quantifies how good the segmentations are compared to real segmentations of the brain. 
To validate the model the compare their 2D VAE with a 3D VAE and other 3D models (3D WGAN GP, 3D VAE GAN, 3D $\alpha$-GAN and 3D $\alpha$-WGAN), finding that their method is able to succesfully sample consistent brain volumes. Even though both their model and the 3D model produce blurry images, that is expected due to the nature of the VAE model. Additionally, they compute the MMD (distance to the original distribution) and the MS-SSIM, showing that their model produces volumes that are both diverse and close to the real distribution of the data. Since these metrics have no anatomical meaning, they also use the Realistic Atlas Score (RAS) metric, finding that the proposed model outperforms state-of-the-art approaches.

On the other hand, the authors in \cite{zhu2023make} employ a latent diffusion approach to generate 3D brain MRI from 2D slices. They argue that generating volumes from 2D MRI often leads to inconsistencies due to the inherent complexity of the data. To address this, they propose an efficient, high-resolution method for generating MRI volumes from 2D images. Their approach utilizes an encoder to generate two latent representations: $z_c$ from the source modality and $z$ from the target modality. These latent spaces are concatenated and then fed into the U-Net of the diffusion process. To extend the model from slice-based to volume-based generation, they incorporate volumetric layers and fine-tune the U-Net. These layers enable the model to learn latent features that account for volumetric coherence across slices, and are implemented using 1D convolutions.

The authors evaluate their model using SSIM, MAE, and PSNR metrics, comparing it against baseline methods such as Pix2Pix, Palette \cite{saharia2022palette}, Pix2Pix 3D, and CycleGAN 3D. Their results show that their model outperforms all others in every metric, across both the S2M and RIRE \cite{west1997comparison} datasets. Qualitative analysis further demonstrates that the proposed model produces synthetic images with higher detail and fewer inconsistencies compared to the baseline models. Additionally, an ablation study reveals that fine-tuning the volumetric layers helps reduce volumetric artifacts. 

Authors in \cite{manasseh2024neurocognitive} propose a model to enhance the interpretability of deep learning models applied to MRI classification. The key idea is to organize the latent space according to clinically meaningful variables, specifically the Neuropsychological Z-global score (NPZ), which measures cognitive impairment. This ensures that differences in the latent space reflect changes in the NPZ score. To achieve this, the latent space is disentangled along a linear direction, $\tau$, that captures variations in NPZ. The model is trained on 3D MRIs and their corresponding NPZ scores $\sigma$, optimizing both a classification loss and a disentanglement loss. The disentanglement loss is defined as:
\begin{equation}
    L_r = \sum_{(i,j)\in D} (||\Delta\tau(i,j)||^2 - |\Delta\sigma(i,j)|),
\end{equation}
where $||\Delta\tau(i,j)||$ represents the difference between the projections of latent representations along $\tau$, and $|\Delta\sigma(i,j)|$ is the difference in NPZ scores for the corresponding MRI pairs. This setup ensures a linear relationship between changes in NPZ and the latent space.

Applied to a diverse dataset of 519 cross-sectional MRIs spanning multiple conditions (MCI, HAND, controls, HIV without HAND), the proposed model achieved a balanced accuracy of $46.13\%$, significantly outperforming the baseline model ($39.9\%$, $p=0.009$). Notably, the disentangled latent space was not only interpretable but also demonstrated robustness across demographic factors such as age, sex, and education, which are often confounding variables in neuroimaging studies.

Saliency maps derived from the model highlighted key brain regions driving classification, such as the superior temporal gyrus and caudate nucleus for MCI, and the cerebellar white matter for HAND. These findings align with known patterns of neurodegeneration reported in the literature, reinforcing the clinical and neuroscientific relevance of the model.

In \cite{li2020latent}, the authors propose a Variational Autoencoder (VAE) framework to derive latent representations from multichannel EEG data, aimed at recognizing emotional states in subjects. The VAE is chosen for its ability to model the underlying probability distributions of the data, making it particularly effective for capturing the intrinsic, emotion-related factors from EEG signals. Recognizing the dynamic and time-dependent nature of emotions and brain activity, the authors integrate Long Short-Term Memory (LSTM) networks to process temporal sequences of the extracted latent factors and identify patterns linked to emotional states.

The framework's performance is evaluated on the DEAP and SEED datasets, using the F1-score as a metric. Results show that the combination of VAE and LSTM significantly outperforms traditional approaches such as ICA and Autoencoder (AE), both in terms of reconstruction and emotion classification. Notably, the VAE demonstrates superior performance with fewer latent dimensions, indicating its efficacy in extracting meaningful latent features from EEG data. This work establishes a robust method for emotion recognition and highlights the potential of generative models in decoding latent cognitive processes.[REVISAR]

In \cite{zimmerer2019unsupervised}, the authors emphasize the need for an assumption-free model capable of detecting anomalies in brain images. They critique current reconstruction-based approaches, which require tailoring the network architecture to the specific image modality and type of anomaly being evaluated. This dependency not only limits generalization but also contradicts the goal of creating flexible and broadly applicable models for medical imaging.

To address this, they propose enhancing the traditional reconstruction term with the Kullback-Leibler (KL) divergence, aiming to improve the robustness and accuracy of anomaly localization. They demonstrate experimentally that relying solely on reconstruction error is suboptimal, as it can lead to limited performance depending on the latent space configuration. On the HCP dataset, they show that models combining reconstruction and KL divergence generally outperform reconstruction-only approaches in anomaly detection, particularly in cases with diverse hyperparameter settings. Even in the few cases where reconstruction error performs better, such performance is achieved under restrictive conditions, such as using a small latent space. Additionally, on the BraTS2017 dataset, their combined method consistently outperforms reconstruction-only methods, further validating its efficacy across datasets with different characteristics.

\section{Discussion}

The use of latent representation models in neuroimaging has demonstrated significant potential fro addressing key challenges in the field, such as the high dimensionality of data, inter-subject variability, and the integration of multimodal datasets. Throughout this review, we have explored the theoretical foundations of latent spaces, their alignment with the manifold hypothesis, and their application across a spectrum of neuroimaging tasks.

Generative models, particularly VAEs, GANs, and LDMs, have proven to be essential tools for uncovering meaningful latent structures in neuroimaging data. VAEs offer probabilistic insights that capture variability in pathological processes, while LDMs excel in generating high-quality synthetic data. GANs, on the other hand, have been instrumental in applications requiring detailed image synthesis and harmonization. Despite these advances, we should also highlight the limitations inherent to each model, including trade-offs between reconstruction quality and latent space interpretability, as well as the challenges posed by data heterogeneity.

From a practical standpoint, the ability of latent generative models to harmonize datasets, reconstruct brain activity, and model disease progression represents a critical step forward advancing personalized medicine and enhancing diagnostic accuracy. For instance, harmonization techniques reduce inter-set variability, enabling robust multi-center studies, while brain reconstruction and ageing models provide new avenues for studying disease progression and neurodevelopment. Studies have shown that LDM and GAN based models stand out among latent models in synthesizing high quality images that encode relevant semantic information \cite{gu2022decoding,takagi2023high}, or harmonization tasks \cite{wang2023spatial, qu2019wavelet} . On the other hand, explicit VAE-based models have proven to outperform most models in capturing patterns of brain progression or neurodegeneration \cite{choi2018predicting, basu2019early, matsubara2019deep, martinez2024bridging}.

Additionally, the role of latent generative models in advancing our understanding of brain function and active inference underscores their potential to bridge computational neuroscience and clinical applications. Inference based model has proven to be a hot topic in the understanding of cognitive processes and unraveling the arise of consciousness \cite{friston2001generative} . Moreover, these models also seem to be a promising research gateway for understanding neural coding and decoding, as we have seen that some studies demonstrate the similarity between latent generative models and visual encoding or delusions formation in the brain \cite{friston2003learning, friston2005theory, gershman2019generative}. 

However, there remain important open questions about the interpretability of latent spaces, the integration of multimodal data, and the scalability of these approaches in clinical settings. For instance, 3T to 7T MRI translation faces the problem of the disentangling latent space, in which the capture of a real anatomical representation and its interpretablity is still an open issue. Moreover, most studies in the literature focus on using the latent representations of neuroimaging as a tool for synthesizing or reconstructing images or volumes, while very little works make use of the information of the latent space to perform statistical analysis. This issue becomes clear as most of the studies use GAN or LDM based-models, which are implicit frameworks. Although these models take into account the learned features of the representations, we have no access to the latent variables, which greatly hinders the analysis and interpretability of the models. Even among the models that use explicit models like VAE, only a few of them, such as \cite{martinez2024bridging, geenjaar2021fusing}, perform statistical methods to the latent variables.

Future research should focus not only on improving these preexisting models but also exploing the exploitability of these representations as a mean for obtaining relevant insights that can be related to brain processes or biomarkers that characterize neurodegenerative diseases. Furthermore, latent generative models based on bayesian inference have proven to excel in furthering our understanding about underlying brain processes, their limitations brings to light the need to pursue new inference frameworks, like empirical bayesian inference \cite{morris1983parametric}.

\renewcommand{\arraystretch}{1.3} 

\begin{longtable}{| p{.25\textwidth} | p{.25\textwidth} | p{.30\textwidth} | p{.05\textwidth} |}

\hline
\textbf{Topic} & \textbf{Architecture} & \textbf{Database} & \textbf{Ref} \\
\hline

\multirow{3}{*}{\makecell[l]{\textbf{Brain as}\\\textbf{an inference}\\\textbf{machine}}} 
 &  & DNN & \cite{friston2001generative} \\
 &  & BI & \cite{friston2010computational} \\
 &  & GAN & \cite{gershman2019generative} \\
\hline
\hline

\multirow{2}{*}{\makecell[l]{\textbf{Image-to-Image}\\\textbf{translation}\\\textbf{\& Harmonization}}}
 & BraTs2015, Iseg2017, MRBrain13, ADNI, RIRE & CGAN & \cite{yang2020mri} \\
 & Private & GAN+non-adversarial losses & \cite{armanious2020medgan}\\
 & BraTS2021, IXI  & MS-SPADE & \cite{kim2024adaptive}\\
 & ADNI, MRI-GENIE & SIT-cGAN & \cite{wang2023spatial}\\
 & OpenBHB, SRPBS & DLEST & \cite{wu2024disentangled}\\
 & OASIS, SRPBS & VAE-GAN & \cite{cackowski2023imunity}\\
 & IXI, OASIS3, BLSA & cVAE+IB loss & \cite{zuo2021unsupervised}\\
 & IXI, 2 Private & U-Net & \cite{dewey2020disentangled}\\
 & Private & GAN + non-adversarial losses & \cite{qu2019wavelet}\\
 
\hline
\hline

\multirow{2}{*}{\makecell[l]{\textbf{Visual}\\\textbf{reconstruction}\\\textbf{using fMRI}}}
 & BRAINS, Generic Object Decoding & GAN+linear regression & \cite{seeliger2018generative}\\
 & ILSVRC2012, Private & VAE+linear regression & \cite{han2019variational} \\
 & NSD & AE+LDM & \cite{takagi2023high} \\
 & NSD & spherical CNN+variational+IC-GAN & \cite{gu2022decoding} \\
 & NSD & VDVAE+regression+LDM & \cite{ozcelik2023natural} \\
 & vim-1 & fwRF+GAN & \cite{st2018generative} \\
 & HCP, NSD & MAE+CLIP+VDVAE+LDM & \cite{guo2024mindldm} \\
\hline
\hline

\multirow{2}{*}{\textbf{Brain ageing analysis}}
 & Cam-CAN, ADNI & VGG+adversarial $\&$ identity losses & \cite{xia2019consistent} \\
 & ADNI & VAE & \cite{choi2018predicting} \\
 & ADNI & VAE+temporal linear model & \cite{chadebec2022image} \\
 & UK Biobank & VAE+ age regression & \cite{calm2024identifying} \\
 & dHCP & INR & \cite{bieder2024modeling} \\
\hline
\hline

\multirow{2}{*}{\textbf{Disease classification}}
 & ADNI & VAE+MLP & \cite{basu2019early} \\
 & ADNI & Negative Matrix Factorization & \cite{zhou2019deep} \\
 & OpenfMRI & CVAE+DGM & \cite{matsubara2019deep} \\
 & Public \cite{deserno2012reduced}  & BI+SVM+GMM & \cite{brodersen2014dissecting} \\
 & ADHD-200 & STAAE+lasso regression & \cite{dong2020spatiotemporal} \\
\hline
\hline

\makecell[l]{\textbf{Functional brain}\\\textbf{networks}}
 & ADHD-200 & DVAE+lasso regression & \cite{qiang2020deep} \\
 & WU-Minn HPC & Convolutional kernel+attention & \cite{zhang2020deep}\\
  & HCP Q3 & 3D ResAE+lasso regression & \cite{dong2020discovering} \\
\hline
\hline

\makecell[l]{\textbf{Multimodality}\\\textbf{integration}}
 & FBIRN, B-SNIP, COBRE & VAE & \cite{geenjaar2021fusing} \\
 & PPMI & joint VAE & \cite{martinez2024bridging} \\
 & Private & Matrix latent representation & \cite{ghosal2019bridging} \\
 & ADNI-1, ADNIGO/2 & AE+cLDM & \cite{jeon2024gene} \\
\hline
\hline

\multirow{2}{*}{\textbf{Image synthesis}}
 & Brain MRI & GAN & \cite{zhang2019skrgan} \\
 & UK Biobank & AE+LDM & \cite{pinaya2022brain} \\
\hline
\hline

\multirow{2}{*}{\textbf{Other works}}
 & HCP & VAE+Gamma model & \cite{volokitin2020modelling} \\
 & SWI-to-MRA, RIRE & LDM+1D convolutions & \cite{zhu2023make} \\
 & UCSF & CNN+disentanglement loss & \cite{manasseh2024neurocognitive} \\
 & DEAP, SEED & VAE+LSTM & \cite{li2020latent} \\
 & HCP, BraTS2017 & VAE & \cite{zimmerer2019unsupervised} \\
\hline

\caption{Overview of the studies.}
\label{tab:review_summary}
\end{longtable}


\bibliographystyle{plain}
\bibliography{refs}

\begin{thebibliography}{10}

\bibitem{armanious2020medgan}
Karim Armanious, Chenming Jiang, Marc Fischer, Thomas K{\"u}stner, Tobias Hepp, Konstantin Nikolaou, Sergios Gatidis, and Bin Yang.
\newblock Medgan: Medical image translation using gans.
\newblock {\em Computerized medical imaging and graphics}, 79:101684, 2020.

\bibitem{bahrami20177t}
Khosro Bahrami, Feng Shi, Islem Rekik, Yaozong Gao, and Dinggang Shen.
\newblock 7t-guided super-resolution of 3t mri.
\newblock {\em Medical physics}, 44(5):1661--1677, 2017.

\bibitem{bahrami2016reconstruction}
Khosro Bahrami, Feng Shi, Xiaopeng Zong, Hae~Won Shin, Hongyu An, and Dinggang Shen.
\newblock Reconstruction of 7t-like images from 3t mri.
\newblock {\em IEEE transactions on medical imaging}, 35(9):2085--2097, 2016.

\bibitem{basu2019early}
Sumana Basu, Konrad Wagstyl, Azar Zandifar, Louis Collins, Adriana Romero, and Doina Precup.
\newblock Early prediction of alzheimer’s disease progression using variational autoencoders.
\newblock In {\em Medical Image Computing and Computer Assisted Intervention--MICCAI 2019: 22nd International Conference, Shenzhen, China, October 13--17, 2019, Proceedings, Part IV 22}, pages 205--213. Springer, 2019.

\bibitem{bieder2024modeling}
Florentin Bieder, Paul Friedrich, H{\'e}l{\`e}ne Corbaz, Alicia Durrer, Julia Wolleb, and Philippe C.~Cattin.
\newblock Modeling the neonatal brain development using implicit neural representations.
\newblock In {\em International Workshop on PRedictive Intelligence In MEdicine}, pages 1--11. Springer, 2024.

\bibitem{brodersen2014dissecting}
Kay~H Brodersen, Lorenz Deserno, Florian Schlagenhauf, Zhihao Lin, Will~D Penny, Joachim~M Buhmann, and Klaas~E Stephan.
\newblock Dissecting psychiatric spectrum disorders by generative embedding.
\newblock {\em NeuroImage: Clinical}, 4:98--111, 2014.

\bibitem{bullmore2012economy}
Ed~Bullmore and Olaf Sporns.
\newblock The economy of brain network organization.
\newblock {\em Nature reviews neuroscience}, 13(5):336--349, 2012.

\bibitem{cackowski2023imunity}
Stenzel Cackowski, Emmanuel~L Barbier, Michel Dojat, and Thomas Christen.
\newblock Imunity: a generalizable vae-gan solution for multicenter mr image harmonization.
\newblock {\em Medical Image Analysis}, 88:102799, 2023.

\bibitem{calm2024identifying}
Berta Calm~Salvans, Irene Cumplido~Mayoral, Juan~Domingo Gispert, and Veronica Vilaplana.
\newblock Identifying brain ageing trajectories using variational autoencoders with regression model in neuroimaging data stratified by sex and validated against dementia-related risk factors.
\newblock In {\em International Workshop on PRedictive Intelligence In MEdicine}, pages 149--160. Springer, 2024.

\bibitem{cao2017t}
Bokai Cao, Lifang He, Xiaokai Wei, Mengqi Xing, Philip~S Yu, Heide Klumpp, and Alex~D Leow.
\newblock t-bne: Tensor-based brain network embedding.
\newblock In {\em proceedings of the 2017 SIAM international conference on data mining}, pages 189--197. SIAM, 2017.

\bibitem{chadebec2022image}
Cl{\'e}ment Chadebec, Evi~MC Huijben, Josien~PW Pluim, St{\'e}phanie Allassonni{\`e}re, and Maureen~AJM van Eijnatten.
\newblock An image feature mapping model for continuous longitudinal data completion and generation of synthetic patient trajectories.
\newblock In {\em MICCAI Workshop on Deep Generative Models}, pages 55--64. Springer, 2022.

\bibitem{choi2018predicting}
Hongyoon Choi, Hyejin Kang, Dong~Soo Lee, and Alzheimer's Disease~Neuroimaging Initiative.
\newblock Predicting aging of brain metabolic topography using variational autoencoder.
\newblock {\em Frontiers in aging neuroscience}, 10:212, 2018.

\bibitem{chung2021neural}
SueYeon Chung and Larry~F Abbott.
\newblock Neural population geometry: An approach for understanding biological and artificial neural networks.
\newblock {\em Current opinion in neurobiology}, 70:137--144, 2021.

\bibitem{coltheart2010abductive}
Max Coltheart, Peter Menzies, and John Sutton.
\newblock Abductive inference and delusional belief.
\newblock {\em Cognitive neuropsychiatry}, 15(1-3):261--287, 2010.

\bibitem{deserno2012reduced}
Lorenz Deserno, Philipp Sterzer, Torsten W{\"u}stenberg, Andreas Heinz, and Florian Schlagenhauf.
\newblock Reduced prefrontal-parietal effective connectivity and working memory deficits in schizophrenia.
\newblock {\em Journal of Neuroscience}, 32(1):12--20, 2012.

\bibitem{dewey2020disentangled}
Blake~E Dewey, Lianrui Zuo, Aaron Carass, Yufan He, Yihao Liu, Ellen~M Mowry, Scott Newsome, Jiwon Oh, Peter~A Calabresi, and Jerry~L Prince.
\newblock A disentangled latent space for cross-site mri harmonization.
\newblock In {\em International conference on medical image computing and computer-assisted intervention}, pages 720--729. Springer, 2020.

\bibitem{dong2020discovering}
Qinglin Dong, Ning Qiang, Jinglei Lv, Xiang Li, Tianming Liu, and Quanzheng Li.
\newblock Discovering functional brain networks with 3d residual autoencoder (resae).
\newblock In {\em Medical Image Computing and Computer Assisted Intervention--MICCAI 2020: 23rd International Conference, Lima, Peru, October 4--8, 2020, Proceedings, Part VII 23}, pages 498--507. Springer, 2020.

\bibitem{dong2020spatiotemporal}
Qinglin Dong, Ning Qiang, Jinglei Lv, Xiang Li, Tianming Liu, and Quanzheng Li.
\newblock Spatiotemporal attention autoencoder (staae) for adhd classification.
\newblock In {\em Medical Image Computing and Computer Assisted Intervention--MICCAI 2020: 23rd International Conference, Lima, Peru, October 4--8, 2020, Proceedings, Part VII 23}, pages 508--517. Springer, 2020.

\bibitem{dyrba2015multimodal}
Martin Dyrba, Michel Grothe, Thomas Kirste, and Stefan~J Teipel.
\newblock Multimodal analysis of functional and structural disconnection in a lzheimer's disease using multiple kernel svm.
\newblock {\em Human brain mapping}, 36(6):2118--2131, 2015.

\bibitem{friedman1997bias}
Jerome~H Friedman.
\newblock On bias, variance, 0/1—loss, and the curse-of-dimensionality.
\newblock {\em Data mining and knowledge discovery}, 1:55--77, 1997.

\bibitem{friston2003learning}
Karl Friston.
\newblock Learning and inference in the brain.
\newblock {\em Neural Networks}, 16(9):1325--1352, 2003.

\bibitem{friston2005theory}
Karl Friston.
\newblock A theory of cortical responses.
\newblock {\em Philosophical transactions of the Royal Society B: Biological sciences}, 360(1456):815--836, 2005.

\bibitem{friston2010computational}
Karl~J Friston and Raymond~J Dolan.
\newblock Computational and dynamic models in neuroimaging.
\newblock {\em Neuroimage}, 52(3):752--765, 2010.

\bibitem{friston2001generative}
Karl~J Friston and Cathy~J Price.
\newblock Generative models, brain function and neuroimaging.
\newblock {\em Scandinavian Journal of Psychology}, 42(3):167--177, 2001.

\bibitem{geenjaar2021fusing}
Eloy Geenjaar, Noah Lewis, Zening Fu, Rohan Venkatdas, Sergey Plis, and Vince Calhoun.
\newblock Fusing multimodal neuroimaging data with a variational autoencoder.
\newblock In {\em 2021 43rd Annual International Conference of the IEEE Engineering in Medicine \& Biology Society (EMBC)}, pages 3630--3633. IEEE, 2021.

\bibitem{gershman2019generative}
Samuel~J Gershman.
\newblock The generative adversarial brain.
\newblock {\em Frontiers in Artificial Intelligence}, 2:18, 2019.

\bibitem{ghosal2019bridging}
Sayan Ghosal, Qiang Chen, Aaron~L Goldman, William Ulrich, Karen~F Berman, Daniel~R Weinberger, Venkata~S Mattay, and Archana Venkataraman.
\newblock Bridging imaging, genetics, and diagnosis in a coupled low-dimensional framework.
\newblock In {\em Medical Image Computing and Computer Assisted Intervention--MICCAI 2019: 22nd International Conference, Shenzhen, China, October 13--17, 2019, Proceedings, Part IV 22}, pages 647--655. Springer, 2019.

\bibitem{goodfellow2014generative}
Ian Goodfellow, Jean Pouget-Abadie, Mehdi Mirza, Bing Xu, David Warde-Farley, Sherjil Ozair, Aaron Courville, and Yoshua Bengio.
\newblock Generative adversarial nets.
\newblock {\em Advances in neural information processing systems}, 27, 2014.

\bibitem{gu2022decoding}
Zijin Gu, Keith Jamison, Amy Kuceyeski, and Mert Sabuncu.
\newblock Decoding natural image stimuli from fmri data with a surface-based convolutional network.
\newblock {\em arXiv preprint arXiv:2212.02409}, 2022.

\bibitem{guo2024mindldm}
Junhao Guo, Chanlin Yi, Fali Li, Peng Xu, and Yin Tian.
\newblock Mindldm: Reconstruct visual stimuli from fmri using latent diffusion model.
\newblock In {\em 2024 IEEE International Conference on Computational Intelligence and Virtual Environments for Measurement Systems and Applications (CIVEMSA)}, pages 1--6. IEEE, 2024.

\bibitem{han2019variational}
Kuan Han, Haiguang Wen, Junxing Shi, Kun-Han Lu, Yizhen Zhang, Di~Fu, and Zhongming Liu.
\newblock Variational autoencoder: An unsupervised model for encoding and decoding fmri activity in visual cortex.
\newblock {\em NeuroImage}, 198:125--136, 2019.

\bibitem{jazayeri2021interpreting}
Mehrdad Jazayeri and Srdjan Ostojic.
\newblock Interpreting neural computations by examining intrinsic and embedding dimensionality of neural activity.
\newblock {\em Current opinion in neurobiology}, 70:113--120, 2021.

\bibitem{jeon2024gene}
Sooyeon Jeon, Yujee Song, and Won~Hwa Kim.
\newblock Gene-to-image: Decoding brain images from genetics via latent diffusion models.
\newblock In {\em International Workshop on PRedictive Intelligence In MEdicine}, pages 48--60. Springer, 2024.

\bibitem{jog2015mr}
Amod Jog, Aaron Carass, Snehashis Roy, Dzung~L Pham, and Jerry~L Prince.
\newblock Mr image synthesis by contrast learning on neighborhood ensembles.
\newblock {\em Medical image analysis}, 24(1):63--76, 2015.

\bibitem{kao2021optimal}
Ta-Chu Kao, Mahdieh~S Sadabadi, and Guillaume Hennequin.
\newblock Optimal anticipatory control as a theory of motor preparation: A thalamo-cortical circuit model.
\newblock {\em Neuron}, 109(9):1567--1581, 2021.

\bibitem{kawahara2017brainnetcnn}
Jeremy Kawahara, Colin~J Brown, Steven~P Miller, Brian~G Booth, Vann Chau, Ruth~E Grunau, Jill~G Zwicker, and Ghassan Hamarneh.
\newblock Brainnetcnn: Convolutional neural networks for brain networks; towards predicting neurodevelopment.
\newblock {\em NeuroImage}, 146:1038--1049, 2017.

\bibitem{kensinger2006neural}
Elizabeth~A Kensinger and Daniel~L Schacter.
\newblock Neural processes underlying memory attribution on a reality-monitoring task.
\newblock {\em Cerebral Cortex}, 16(8):1126--1133, 2006.

\bibitem{kim2024adaptive}
Jonghun Kim and Hyunjin Park.
\newblock Adaptive latent diffusion model for 3d medical image to image translation: Multi-modal magnetic resonance imaging study.
\newblock In {\em Proceedings of the IEEE/CVF Winter Conference on Applications of Computer Vision}, pages 7604--7613, 2024.

\bibitem{kingma2013auto}
Diederik~P Kingma.
\newblock Auto-encoding variational bayes.
\newblock {\em arXiv preprint arXiv:1312.6114}, 2013.

\bibitem{ktena2018metric}
Sofia~Ira Ktena, Sarah Parisot, Enzo Ferrante, Martin Rajchl, Matthew Lee, Ben Glocker, and Daniel Rueckert.
\newblock Metric learning with spectral graph convolutions on brain connectivity networks.
\newblock {\em NeuroImage}, 169:431--442, 2018.

\bibitem{larsen2016autoencoding}
Anders Boesen~Lindbo Larsen, S{\o}ren~Kaae S{\o}nderby, Hugo Larochelle, and Ole Winther.
\newblock Autoencoding beyond pixels using a learned similarity metric.
\newblock In {\em International conference on machine learning}, pages 1558--1566. PMLR, 2016.

\bibitem{li2020latent}
Xiang Li, Zhigang Zhao, Dawei Song, Yazhou Zhang, Jingshan Pan, Lu~Wu, Jidong Huo, Chunyang Niu, and Di~Wang.
\newblock Latent factor decoding of multi-channel eeg for emotion recognition through autoencoder-like neural networks.
\newblock {\em Frontiers in neuroscience}, 14:87, 2020.

\bibitem{lindsay2022uncovering}
Grace Lindsay.
\newblock {\em Uncovering Hidden Dimensions in Brain Signals}.
\newblock MIT Press, Cambridge, MA, 2022.

\bibitem{lu2022image}
Jiahao Lu, Johan {\"O}fverstedt, Joakim Lindblad, and Nata{\v{s}}a Sladoje.
\newblock Is image-to-image translation the panacea for multimodal image registration? a comparative study.
\newblock {\em Plos one}, 17(11):e0276196, 2022.

\bibitem{manasseh2024neurocognitive}
Jocasta Manasseh-Lewis, Felipe Godoy, Wei Peng, Robert Paul, Ehsan Adeli, and Kilian Pohl.
\newblock Neurocognitive latent space regularization for multi-label diagnosis from mri.
\newblock In {\em International Workshop on PRedictive Intelligence In MEdicine}, pages 185--195. Springer, 2024.

\bibitem{mao2017least}
Xudong Mao, Qing Li, Haoran Xie, Raymond~YK Lau, Zhen Wang, and Stephen Paul~Smolley.
\newblock Least squares generative adversarial networks.
\newblock In {\em Proceedings of the IEEE international conference on computer vision}, pages 2794--2802, 2017.

\bibitem{martinez2024bridging}
Francisco~J Martinez-Murcia, Juan~Eloy Arco, Carmen Jimenez-Mesa, Fermin Segovia, Ignacio~A Illan, Javier Ramirez, and Juan~Manuel Gorriz.
\newblock Bridging imaging and clinical scores in parkinson's progression via multimodal self-supervised deep learning.
\newblock {\em International Journal of Neural Systems}, pages 2450043--2450043, 2024.

\bibitem{matsubara2019deep}
Takashi Matsubara, Tetsuo Tashiro, and Kuniaki Uehara.
\newblock Deep neural generative model of functional mri images for psychiatric disorder diagnosis.
\newblock {\em IEEE Transactions on Biomedical Engineering}, 66(10):2768--2779, 2019.

\bibitem{morris1983parametric}
Carl~N Morris.
\newblock Parametric empirical bayes inference: theory and applications.
\newblock {\em Journal of the American statistical Association}, 78(381):47--55, 1983.

\bibitem{navab2015medical}
Nassir Navab, Joachim Hornegger, William~M Wells, and Alejandro Frangi.
\newblock {\em Medical Image Computing and Computer-Assisted Intervention--MICCAI 2015: 18th International Conference, Munich, Germany, October 5-9, 2015, Proceedings, Part III}, volume 9351.
\newblock Springer, 2015.

\bibitem{ozcelik2023natural}
Furkan Ozcelik and Rufin VanRullen.
\newblock Natural scene reconstruction from fmri signals using generative latent diffusion.
\newblock {\em Scientific Reports}, 13(1):15666, 2023.

\bibitem{pinaya2022dataset}
Walter H.~L. Pinaya, Petru-Daniel Tudosiu, Jessica Dafflon, Pedro F.~Da Costa, Virginia Fernandez, Parashkev Nachev, Sébastien Ourselin, and M.~Jorge Cardoso.
\newblock Synthetic dataset of 100,000 brain mri scans, 2022.
\newblock Available at Academic Torrents.

\bibitem{pinaya2022brain}
Walter~HL Pinaya, Petru-Daniel Tudosiu, Jessica Dafflon, Pedro~F Da~Costa, Virginia Fernandez, Parashkev Nachev, Sebastien Ourselin, and M~Jorge Cardoso.
\newblock Brain imaging generation with latent diffusion models.
\newblock In {\em MICCAI Workshop on Deep Generative Models}, pages 117--126. Springer, 2022.

\bibitem{premack1978does}
David Premack and Guy Woodruff.
\newblock Does the chimpanzee have a theory of mind?
\newblock {\em Behavioral and brain sciences}, 1(4):515--526, 1978.

\bibitem{qiang2020deep}
Ning Qiang, Qinglin Dong, Fangfei Ge, Hongtao Liang, Bao Ge, Shu Zhang, Yifei Sun, Jie Gao, and Tianming Liu.
\newblock Deep variational autoencoder for mapping functional brain networks.
\newblock {\em IEEE Transactions on Cognitive and Developmental Systems}, 13(4):841--852, 2020.

\bibitem{qu2019wavelet}
Liangqiong Qu, Shuai Wang, Pew-Thian Yap, and Dinggang Shen.
\newblock Wavelet-based semi-supervised adversarial learning for synthesizing realistic 7t from 3t mri.
\newblock In {\em Medical Image Computing and Computer Assisted Intervention--MICCAI 2019: 22nd International Conference, Shenzhen, China, October 13--17, 2019, Proceedings, Part IV 22}, pages 786--794. Springer, 2019.

\bibitem{radford2021clip}
Alec Radford, Jongwei Kim, C.~Hallacy, Aditya Ramesh, Gabriel Goh, Shibani Agarwal, Gauri Sastry, Amanda Askell, Paul Mishkin, Jack Clark, and et~al.
\newblock Learning transferable visual models from natural language supervision.
\newblock {\em arXiv preprint arXiv:2103.00020}, 2021.

\bibitem{ramachandran1997three}
Vilayanur~S Ramachandran and William Hirstein.
\newblock Three laws of qualia: What neurology tells us about the biological functions of consciousness.
\newblock {\em Journal of consciousness studies}, 4(5-6):429--457, 1997.

\bibitem{roweis2000nonlinear}
Sam~T Roweis and Lawrence~K Saul.
\newblock Nonlinear dimensionality reduction by locally linear embedding.
\newblock {\em science}, 290(5500):2323--2326, 2000.

\bibitem{saharia2022palette}
Chitwan Saharia, William Chan, Huiwen Chang, Chris Lee, Jonathan Ho, Tim Salimans, David Fleet, and Mohammad Norouzi.
\newblock Palette: Image-to-image diffusion models.
\newblock In {\em ACM SIGGRAPH 2022 conference proceedings}, pages 1--10, 2022.

\bibitem{seeliger2018generative}
Katja Seeliger, Umut G{\"u}{\c{c}}l{\"u}, Luca Ambrogioni, Yagmur G{\"u}{\c{c}}l{\"u}t{\"u}rk, and Marcel~AJ Van~Gerven.
\newblock Generative adversarial networks for reconstructing natural images from brain activity.
\newblock {\em NeuroImage}, 181:775--785, 2018.

\bibitem{st2018feature}
Ghislain St-Yves and Thomas Naselaris.
\newblock The feature-weighted receptive field: an interpretable encoding model for complex feature spaces.
\newblock {\em NeuroImage}, 180:188--202, 2018.

\bibitem{st2018generative}
Ghislain St-Yves and Thomas Naselaris.
\newblock Generative adversarial networks conditioned on brain activity reconstruct seen images.
\newblock In {\em 2018 IEEE international conference on systems, man, and cybernetics (SMC)}, pages 1054--1061. IEEE, 2018.

\bibitem{sui2011discriminating}
Jing Sui, Godfrey Pearlson, Arvind Caprihan, T{\"u}lay Adali, Kent~A Kiehl, Jingyu Liu, Jeremy Yamamoto, and Vince~D Calhoun.
\newblock Discriminating schizophrenia and bipolar disorder by fusing fmri and dti in a multimodal cca+ joint ica model.
\newblock {\em Neuroimage}, 57(3):839--855, 2011.

\bibitem{takagi2023high}
Yu~Takagi and Shinji Nishimoto.
\newblock High-resolution image reconstruction with latent diffusion models from human brain activity.
\newblock In {\em Proceedings of the IEEE/CVF Conference on Computer Vision and Pattern Recognition}, pages 14453--14463, 2023.

\bibitem{tenenbaum2000global}
Joshua~B Tenenbaum, Vin~de Silva, and John~C Langford.
\newblock A global geometric framework for nonlinear dimensionality reduction.
\newblock {\em science}, 290(5500):2319--2323, 2000.

\bibitem{vaswani2017attention}
Ashish Vaswani, Noam Shazeer, Niki Parmar, Jakob Uszkoreit, Llion Jones, Aidan~N. Gomez, Łukasz Kaiser, and Illia Polosukhin.
\newblock Attention is all you need.
\newblock In {\em Advances in Neural Information Processing Systems (NeurIPS)}, pages 5998--6008, 2017.

\bibitem{volokitin2020modelling}
Anna Volokitin, Ertunc Erdil, Neerav Karani, Kerem~Can Tezcan, Xiaoran Chen, Luc Van~Gool, and Ender Konukoglu.
\newblock Modelling the distribution of 3d brain mri using a 2d slice vae.
\newblock In {\em Medical Image Computing and Computer Assisted Intervention--MICCAI 2020: 23rd International Conference, Lima, Peru, October 4--8, 2020, Proceedings, Part VII 23}, pages 657--666. Springer, 2020.

\bibitem{wang2023spatial}
Clinton~J Wang, Natalia~S Rost, and Polina Golland.
\newblock Spatial-intensity transforms for medical image-to-image translation.
\newblock {\em IEEE transactions on medical imaging}, 42(11):3362--3373, 2023.

\bibitem{west1997comparison}
Jay West, J~Michael Fitzpatrick, Matthew~Y Wang, Benoit~M Dawant, Calvin~R Maurer~Jr, Robert~M Kessler, Robert~J Maciunas, Christian Barillot, Didier Lemoine, Andre Collignon, et~al.
\newblock Comparison and evaluation of retrospective intermodality brain image registration techniques.
\newblock {\em Journal of computer assisted tomography}, 21(4):554--568, 1997.

\bibitem{wu2024disentangled}
Mengqi Wu, Lintao Zhang, Pew-Thian Yap, Hongtu Zhu, and Mingxia Liu.
\newblock Disentangled latent energy-based style translation: An image-level structural mri harmonization framework.
\newblock {\em arXiv preprint arXiv:2402.06875}, 2024.

\bibitem{xia2019consistent}
Tian Xia, Agisilaos Chartsias, Sotirios~A Tsaftaris, and Alzheimer’s Disease~Neuroimaging Initiative.
\newblock Consistent brain ageing synthesis.
\newblock In {\em Medical Image Computing and Computer Assisted Intervention--MICCAI 2019: 22nd International Conference, Shenzhen, China, October 13--17, 2019, Proceedings, Part IV 22}, pages 750--758. Springer, 2019.

\bibitem{yang2020mri}
Qianye Yang, Nannan Li, Zixu Zhao, Xingyu Fan, Eric I-Chao Chang, and Yan Xu.
\newblock Mri cross-modality image-to-image translation.
\newblock {\em Scientific reports}, 10(1):3753, 2020.

\bibitem{yoshida2010neural}
Wako Yoshida, Ben Seymour, Karl~J Friston, and Raymond~J Dolan.
\newblock Neural mechanisms of belief inference during cooperative games.
\newblock {\em Journal of Neuroscience}, 30(32):10744--10751, 2010.

\bibitem{zhang2019skrgan}
Tianyang Zhang, Huazhu Fu, Yitian Zhao, Jun Cheng, Mengjie Guo, Zaiwang Gu, Bing Yang, Yuting Xiao, Shenghua Gao, and Jiang Liu.
\newblock Skrgan: Sketching-rendering unconditional generative adversarial networks for medical image synthesis.
\newblock In {\em Medical Image Computing and Computer Assisted Intervention--MICCAI 2019: 22nd International Conference, Shenzhen, China, October 13--17, 2019, Proceedings, Part IV 22}, pages 777--785. Springer, 2019.

\bibitem{zhang2020deep}
Wen Zhang, Liang Zhan, Paul Thompson, and Yalin Wang.
\newblock Deep representation learning for multimodal brain networks.
\newblock In {\em Medical Image Computing and Computer Assisted Intervention--MICCAI 2020: 23rd International Conference, Lima, Peru, October 4--8, 2020, Proceedings, Part VII 23}, pages 613--624. Springer, 2020.

\bibitem{zhang2018dual}
Yongqin Zhang, Jie-Zhi Cheng, Lei Xiang, Pew-Thian Yap, and Dinggang Shen.
\newblock Dual-domain cascaded regression for synthesizing 7t from 3t mri.
\newblock In {\em Medical Image Computing and Computer Assisted Intervention--MICCAI 2018: 21st International Conference, Granada, Spain, September 16-20, 2018, Proceedings, Part I}, pages 410--417. Springer, 2018.

\bibitem{zhou2019deep}
Tao Zhou, Mingxia Liu, Huazhu Fu, Jun Wang, Jianbing Shen, Ling Shao, and Dinggang Shen.
\newblock Deep multi-modal latent representation learning for automated dementia diagnosis.
\newblock In {\em International conference on medical image computing and computer-assisted intervention}, pages 629--638. Springer, 2019.

\bibitem{zhu2023make}
Lingting Zhu, Zeyue Xue, Zhenchao Jin, Xian Liu, Jingzhen He, Ziwei Liu, and Lequan Yu.
\newblock Make-a-volume: Leveraging latent diffusion models for cross-modality 3d brain mri synthesis.
\newblock In {\em International Conference on Medical Image Computing and Computer-Assisted Intervention}, pages 592--601. Springer, 2023.

\bibitem{zimmerer2019unsupervised}
David Zimmerer, Fabian Isensee, Jens Petersen, Simon Kohl, and Klaus Maier-Hein.
\newblock Unsupervised anomaly localization using variational auto-encoders.
\newblock In {\em Medical Image Computing and Computer Assisted Intervention--MICCAI 2019: 22nd International Conference, Shenzhen, China, October 13--17, 2019, Proceedings, Part IV 22}, pages 289--297. Springer, 2019.

\bibitem{zuo2021unsupervised}
Lianrui Zuo, Blake~E Dewey, Yihao Liu, Yufan He, Scott~D Newsome, Ellen~M Mowry, Susan~M Resnick, Jerry~L Prince, and Aaron Carass.
\newblock Unsupervised mr harmonization by learning disentangled representations using information bottleneck theory.
\newblock {\em NeuroImage}, 243:118569, 2021.

\end{thebibliography}

\end{document}